\pdfoutput=1
\documentclass[11pt]{article}

\usepackage[]{ACL2023}

\usepackage{times}
\usepackage{latexsym}
\usepackage{amsmath}
\usepackage{amssymb}
\usepackage{kotex}
\usepackage[toc,page]{appendix}
\usepackage{multirow}
\usepackage{caption}
\usepackage{longtable}
\usepackage{enumitem}% http://ctan.org/pkg/enumitem
\usepackage{refcount}
\usepackage[normalem]{ulem}

\usepackage[small,compact]{titlesec}
\usepackage[T1]{fontenc}
\usepackage[utf8]{inputenc}

\usepackage{microtype}
\usepackage{gb4e} % Leipzig (linguistic) glossing
\noautomath % turn off redefinition
\usepackage{tipa} % International Phonetic Alphabet (IPA)
\usepackage{multirow}
\usepackage{multicol}
\usepackage{threeparttable, booktabs, caption}
\usepackage{tabularx}
\usepackage{cellspace}
\usepackage[utf8]{inputenc}
\usepackage[english]{babel}

\usepackage{graphicx}
\usepackage[toc,page]{appendix}
\usepackage{makecell}
\usepackage{ulem} % to cross out sentences (취소선)

\usepackage{subfig}

\usepackage{inconsolata}

\newcommand{\klang}{Kor-Lang8}
\newcommand{\knative}{Kor-Native}
\newcommand{\klearner}{Kor-Learner}
\newcommand{\kunion}{Kor-Union}
\newcommand{\mm}{$M^2$}

\newcommand{\sepauthor}{%
    \end{tabular}%
    \begin{tabular}[t]{c}%
}%

\title{{Towards standardizing Korean Grammatical Error Correction: \\ Datasets and Annotation}}

\author{
    \begin{tabular}[t]{c}
    \textbf{\normalfont Soyoung Yoon}\\
    \small{\normalfont KAIST AI}\\
    \scriptsize{\texttt{\normalfont soyoungyoon@kaist.ac.kr}}
    \end{tabular}
    \sepauthor
    Sungjoon Park\\
    \small{SoftlyAI Research, SoftlyAI}\\
    \scriptsize{\texttt{\normalfont sungjoon.park@softly.ai}}
    \sepauthor
    Gyuwan Kim\\
    \small{UCSB\thanks{\quad University of California, Santa Barbara}}\\
    \scriptsize{\texttt{\normalfont gyuwankim@ucsb.edu}}
    \sepauthor
    Junhee Cho\\
    \small{Google}\\
    \scriptsize{\texttt{\normalfont junheecho@google.com}}
    \AND
    \normalfont Kihyo Park\\
    \small{Cornell University}\\
    \scriptsize{\texttt{\normalfont kp467@cornell.edu}}
    \sepauthor
    Gyutae Kim\\
    \small{SoftlyAI}\\
    \scriptsize{\texttt{\normalfont gt.kim@softly.ai}}
    \sepauthor
    Minjoon Seo\\
    \small{KAIST AI}\\
    \scriptsize{\texttt{\normalfont minjoon@kaist.ac.kr}}
    \sepauthor
    Alice Oh\\
    \small{KAIST SC\thanks{\quad School of Computing, KAIST}}\\
    \scriptsize{\texttt{\normalfont alice.oh@kaist.edu}}
}

\begin{document}
\maketitle
\begin{abstract}
Research on Korean grammatical error correction (GEC) is limited, compared to other major languages such as English.
We attribute this problematic circumstance to the lack of a carefully designed evaluation benchmark for Korean GEC.
In this work, we collect three datasets from different sources (\klang, \knative, and \klearner) that covers a wide range of Korean grammatical errors. Considering the nature of Korean grammar, We then define 14 error types for Korean and provide KAGAS (Korean Automatic Grammatical error Annotation System), which can automatically annotate error types from parallel corpora. We use KAGAS on our datasets to make an evaluation benchmark for Korean, and present baseline models trained from our datasets. We show that the model trained with our datasets significantly outperforms the currently used statistical Korean GEC system (Hanspell) on a wider range of error types, demonstrating the diversity and usefulness of the datasets. The implementations and datasets are open-sourced.\footnote{Code for model checkpoints and KAGAS: \url{https://github.com/soyoung97/Standard_Korean_GEC}. Dataset request form: \url{https://forms.gle/kF9pvJbLGvnh8ZnQ6}}

\end{abstract}
\section{Introduction}
Writing grammatically correct Korean sentences is difficult for learners studying Korean as a Foreign Language (KFL) and even for native Korean speakers due to its morphological and orthographical complexity such as particles, spelling, and collocation. 
% According to the National Institute of Korean Language (NIKL), native Koreans study Korean grammar during the compulsory education courses and are able to pass the required standardized test, but they still have difficulty applying it correctly in daily usage.% 
Its word spacing rule is complex since there are many domain-dependent exceptions, of which only around 20\% of native speakers understand thoroughly \cite{OUNHCHR2014}. 
Since Korean is an agglutinative language \cite{sohn2001korean, song2006korean}, getting used to Korean grammar is time-consuming for KFL learners whose mother tongue is non-agglutinative \cite{haupt2017identifying, kim2020kfl}. However, despite the growing number of KFL learners \cite{lee2018effects}, little research has been conducted on Korean Grammatical Error Correction (GEC) because of the previously described difficulties of the Korean language.
Another major obstacle to developing a Korean GEC system is the lack of resources to train a machine learning model. 

In this paper, we propose three datasets that cover various grammatical errors from different types of annotators and learners. The first dataset named \knative{} is crowd-sourced from native Korean speakers. Second, \klearner{} are from KFL learners that consists of essays with detailed corrections and annotations by Korean tutors. Third, \klang{} are similar with \klearner{} except that they consist of sentences made by KFL learners but corrected by native Koreans on social platforms who are not necessarily linguistic experts.
We also analyze our datasets in terms of error type distributions.

While our proposed parallel corpora can be served as a valuable resource to train a machine learning model, another concern is about the annotation of the datasets. Most existing datasets do not have annotation, which makes it hard to use them for evaluation. A major weakness of human annotation is that (1) experts specialized in Korean Grammar are expensive to hire, (2) making them annotate a large number of parallel corpora is not scalable, and (3) the error types and schema are different by datasets and annotators, which is counter-productive.
Another way that we can analyze and evaluate on the dataset is by automatic annotation from parallel corpora. While there is already one for English called ERRANT \cite{bryant-etal-2017-automatic}, there is no automatic error type detection system for Korean. We cannot fully demonstrate and classify error types and edits by using ERRANT, because Korean has many different characteristics than English (Section \ref{sec:about_korean}). This motivates us to develop an automated error correction system for Korean (KAGAS), along with annotated error types of refined corpora using KAGAS.

Lastly, we build a simple yet effective baseline model based on BART \cite{lewis2019bart} trained from our datasets. We further analyze the generated outputs of BART on how the accuracy of each system differs by error types when compared with a statistical method called Hanspell,\footnote{\url{https://speller.cs.pusan.ac.kr/}} providing use cases and insights gained from analysis.
To summarize, the contributions of this paper are as follows: (1) collection of three different types of parallel corpora for Korean GEC, (2) a novel grammatical error annotation toolkit for Korean called KAGAS, and (3) a simple yet effective open-sourced baseline Korean GEC models trained on our datasets with detailed analysis by KAGAS.

\section{Related Work}

\paragraph{Datasets}
Well-curated datasets in each language are crucial to build a GEC system that can capture language-specific characteristics \cite{bender2011achieving}.
In addition to several shared tasks on English GEC \cite{ng-etal-2014-conll, bryant-etal-2019-bea, rao-etal-2018-overview}, resources for GEC in other languages are also available \cite{wu-etal-2018-cyut, li-etal-2018-hybrid, rozovskaya-roth-2019-grammar, koyama-etal-2020-construction, boyd-2018-using}.
Existing works on Korean GEC \cite{min2020grammatical, lee2021korean, park2020self} are challenging to be replicated because they use internal datasets or existing datasets without providing pre-processing details and scripts.
Therefore, it is urgent to provide publicly available datasets in a unified and easily accessible form with pre-processing pipelines that are fully reproducible for the GEC research on Korean.

\paragraph{Evaluation}
\label{related_work:evaluation}
$M^2$ scorer \cite{dahlmeier-ng-2012-better} which measures precision, recall, and $F_{0.5}$ scores based on edits, is the standard evaluation metric for English GEC models.
It requires an $M^2$ file with annotations of edit paths from an erroneous sentence to a corrected sentence.
However, it is expensive to collect the annotations by human workers as they are often required to have expert linguistic knowledge.
When these annotations are not available, GLEU \cite{napoles-etal-2015-ground}, a simple variant of BLEU \cite{papineni-etal-2002-bleu}, is used instead by the simple n-gram matching.
Another way of generating an $M^2$ file for English in a rule-based manner is by using the error annotation toolkit called ERRANT \cite{bryant-etal-2017-automatic}. We extend ERRANT to make KAGAS and utilize it to align and annotate edits on our datasets and make an $M^2$ file to evaluate on Korean GEC models.

\paragraph{Models}
Early works on Korean GEC focus on detecting particle errors with statistical methods \cite{lee-etal-2012-developing, israel-etal-2013-detecting, dickinson-etal-2011-developing}.
A copy-augmented transformer \cite{zhao-etal-2019-improving} by pre-training to denoise and fine-tuning with paired data demonstrates remarkable performance and is widely used in GEC.
Recent studies \cite{min2020grammatical, lee2021korean, park2020self} apply this method for Korean GEC.
On the other hand, \citet{katsumata-komachi-2020-stronger} show that BART \cite{lewis-etal-2020-bart}, known to be effective on conditioned generation tasks, can be used to build a strong baseline for GEC systems.
Following this work, we load the pre-trained weights from KoBART,\footnote{\url{https://github.com/SKT-AI/KoBART}} a Korean version of BART, and finetune it using our GEC datasets.

\section{Data Collection}

\begin{figure*}
\small
\resizebox{\textwidth}{!}{
\includegraphics[width=\linewidth]{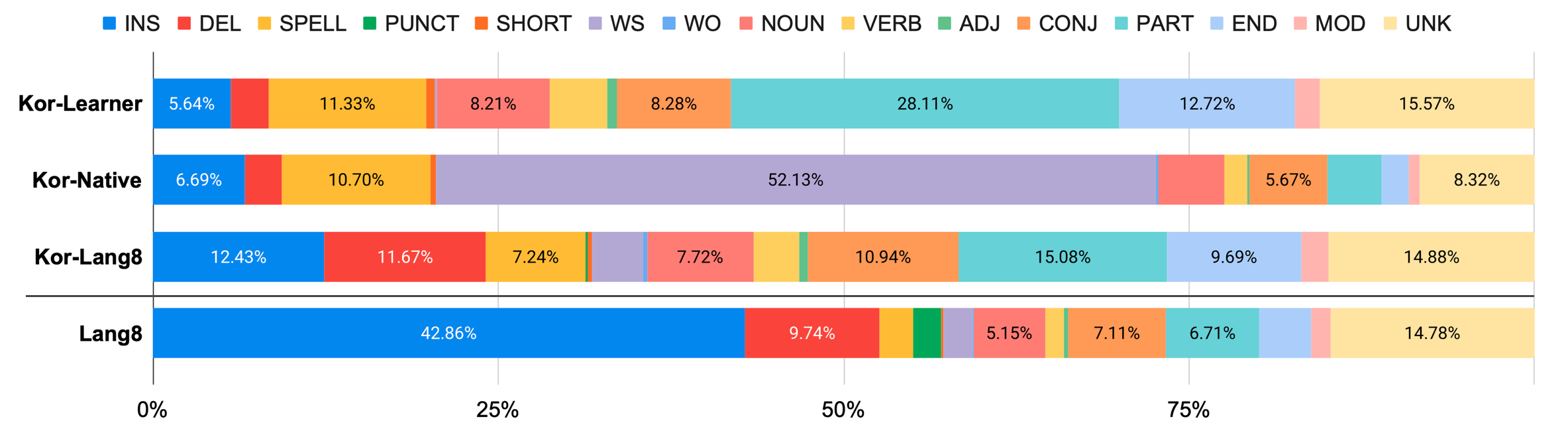}
}
\caption{The distribution of error types by our proposed dataset (Upper three). Percentages are shown for error types that has larger than 5\%. The bottom row (Lang8) is for comparison with \klang{}. We can see that \klang{} has fewer outliers and with decrease in INS or DEL edits than the original Lang8 (bottom). The distribution of error types for KFL learners show that NOUN, CONJ, END and PART are frequent than native learners, and word space errors (purple) are the most frequent for native learners (\knative), similar with previous corpus studies \cite{kim2020kfl, lee2020kor}.
}
\label{fig:errortype_distribution}
\end{figure*}

We build three corpora for Korean GEC: \klearner~(\textsection \ref{sec:klearner}), \knative~(\textsection \ref{sec:knative}), and \klang~(\textsection \ref{sec:klang}).
The statistics of each dataset is described on Table \ref{tab:collection_stats}. We describe the main characteristic and source of the dataset and how it is preprocessed in the following subsection.
We expect that different characteristics of these diverse datasets in terms of quantity, quality, and error type distributions (Figure \ref{fig:errortype_distribution}) allow us to train and evaluate a robust GEC model.

\begin{table}[!t]
\centering
\small
\setlength\tabcolsep{2pt}
\resizebox{0.49\textwidth}{!}{
\begin{tabular}{@{}cccc@{}}
\toprule
 & \textbf{\klearner} & \textbf{\knative} & \textbf{\klang} \\ 
 \midrule
\# Sentence pairs & 28,426 & 17,559 & 109,559 \\ 
\midrule
Avg. token length & 14.86 & 15.22 & 13.07 \\ 
%\midrule
\# Edits  & 59,419 & 29,975 & 262,833 \\ 
% \midrule
\# Edits / sentence & 2.09 & 1.71 & 2.40 \\ 
Avg. tokens per edit & 0.97 & 1.40 & 0.92 \\
Prop. tokens changed & 28.01\% & 29.37\% & 39.42\% \\
\bottomrule
\end{tabular}}
\caption{Data statistics for \klearner{}, \klang{}, and \knative{}.}
\label{tab:collection_stats}
\end{table}

\subsection{Korean Learner Corpus (\klearner)} 
\label{sec:klearner}
Korean Learner Corpus is made from the NIKL learner corpus\footnote{The NIKL learner corpus is created by the National Institute of Korean Language (NIKL). Usage is allowed only for research purposes, and citation of the origin (NIKL) is needed when using it.}. The NIKL learner corpus contains essays written by Korean learners and their grammatical error correction annotations by their tutors in an morpheme-level XML file format. The original format is described at Appendix \ref{appendix:klearner_format}. Even though the NIKL learner corpus contains annotations by professional Korean tutors, it is not possible to directly be used as a corpus for training and evaluation for two reasons.
First, we cannot recover the corrected sentence from the original file nor convert the dataset into an $M^2$ file format (Section \ref{related_work:evaluation}) since the dataset is given by \textit{morpheme-level} (syllable-level) correction annotations, not word-level edits. A simple concatenation of morpheme-level edits does not make a complete word since Korean is an agglutinative language. Therefore, we refer to the current Korean orthography guidelines\footnote{\label{kdnorms}\url{http://kornorms.korean.go.kr/regltn/regltnView.do}} to merge morpheme-level syllables into Korean words (Appendix \ref{appendix:klearner_merging} \footnote{The merging codes are also open-sourced at our repository.}). Second, some XML files had empty edits, missing tags, and inconsistent edit correction tags depending on annotators, so additional refinement and proofreading was required. Therefore, the authors manually inspected the output of parallel corpora and discard sentences with insufficient annotations (Appendix \ref{appendix:klearner_manual}). After applying appropriate modifications to the NIKL corpus, we were able to make \klearner{} which contains word-level parallel sentences with high quality.

\subsection{Native Korean Corpus (\knative)}
\label{sec:knative}
The purpose of this corpus is to build a parallel corpus representing grammatical errors native Korean speakers make.
Because the Korean orthography guidelines are complicated consisting of 57 rules with numerous exceptions,\footnotemark[\getrefnumber{kdnorms}] only a few native Korean speakers fully internalize all from the guidelines and apply them correctly. Thus, the standard approach depends on the manpower of Korean language experts, which is not scalable and is very costly. Thus, we introduce our novel method to create a large parallel GEC corpus from correct sentences, which does not depend on the manpower of experts, but the general public of native Korean speakers. Our method is characterized as a \textit{backward} approach. We collect grammatically correct sentences from two sources,\footnote{(1) The Center for Teaching and Learning for Korean, and (2) National Institute of Korean language} and read the correct sentences using Google Text-to-Speech (TTS) system. We asked the general public to dictate grammatically correct sentences and transcribe them. The transcribed sentences may be incorrect, containing grammatical errors that the audience often makes. Figure \ref{fig:errortype_distribution} shows that most of the collected error types were on word spacing. While the distributions of transcribed and written language cannot be exactly identical, we observe that the error type distribution of \knative{} aligns with that of Native Korean~\citep{shin2015} in that they are dominated by word spacing errors, which means that the types of errors of Kor-Native can serve as a reasonable representative to real-world writing errors made by Native Korean. After the filtering process described in Appendix \ref{appendix:native_collection}, we have 17,559 sentence pairs containing grammatical errors.

\subsection{Lang-8 Korean Corpus (\klang)}
\label{sec:klang}
\begin{table}[t]
\centering
\small

\begin{threeparttable}
\resizebox{0.48\textwidth}{!}{
\begin{tabular}{@{}lllll@{}}
\toprule
 &
  \begin{tabular}[c]{@{}l@{}}Valid \\ loss\end{tabular} &
  \begin{tabular}[c]{@{}l@{}}Self-\\ GLEU\end{tabular} &
  \begin{tabular}[c]{@{}l@{}}GLEU on \\ KoBART\end{tabular} &
  \begin{tabular}[c]{@{}l@{}}Dataset \\ size\end{tabular} \\ \midrule
Lang8 (Bef.) &
  1.53 &
  15.01 &
  19.69 &
  204,130 \\
\klang (Aft.) &
  \textbf{0.83} &
  \textbf{19.38} &
  \textbf{28.57} &
  109,559 \\ \bottomrule
\end{tabular}}
\caption{
\label{datatables:klang}
Evaluation scores on the validation set for Lang8 \cite{mizumoto2011mining}, the original lang8 dataset filtered by unique pairs in Korean, and \klang{}, which is after the refinement by \textsection \ref{sec:klang}.
}
\end{threeparttable}
\end{table}

Lang-8\footnote{\url{https://lang-8.com}} is one of the largest social platforms for language learners \cite{mizumoto2011mining}. 
We extract Korean data from the NAIST Lang-8 Learner Corpora\footnote{\url{https://sites.google.com/site/naistlang8corpora/}} by the language label, resulting in 21,779 Korean sentence pairs.
However, some texts are answers to language-related questions rather than corrections. 
The texts inside the raw Lang-8 corpus is noisy and not all of them form pairs, as previous works with building Japanese corpus out of Lang-8 \cite{koyama-etal-2020-construction} also pointed out.
To build a GEC dataset with high proportion of grammatical edits, we filtered out sentence pairs with a set of cleanup rules regarding the Korean linguistics, which is described in Appendix \ref{appendix:lang8}.

\noindent \textbf{Comparison with original Lang8.}
To prove the increased quality of \klang{}, we compare the model training results and error type distribution between the original Korean version of Lang8 and \klang{}. We perform minimum pre-processing to the original Korean Lang8-data which discard texts that do not have pairs and preserve unique original-corrected sentence pairs to enable training and make a fair comparison with \klang{}, leaving out 204,130 pairs. %We get two key results from the experiment.
Table \ref{datatables:klang} shows that a model trained with \klang{} achieve better results with lower validation loss, higher self-GLEU scores (\textsection \ref{sec:experiments_evaluation}), and higher scores when trained with KoBART, showing that there are \textbf{fewer outliers} on \klang{}.
\footnote{Since the redistribution of the NAIST Lang-8 Learner Corpora is not allowed, we provide the full script used to automatically make \klang{} with the permission of using the corpora for the research purpose only.}
Figure \ref{fig:errortype_distribution} shows the difference in error type distributions before and after Lang8 refinement.

\section{KAGAS}
\label{KAGAS}
\begin{figure}[!ht]
{
\includegraphics[width=\linewidth]{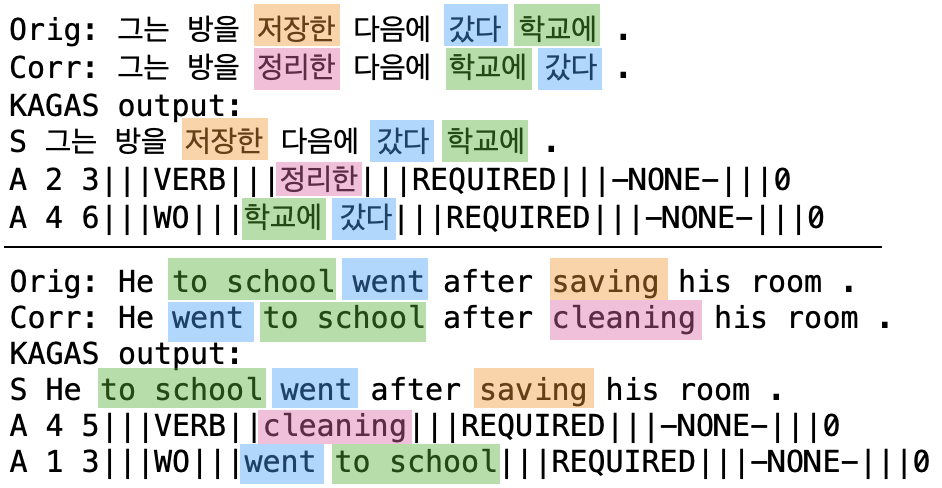}
}
\caption{An example of an $M^2$ file output by KAGAS. Translated into English. Note that "to school" is treated as one word for the translation example. }
\label{fig:kagas_examples}
\end{figure}

\begin{table*}[ht]
\centering
\resizebox{0.8\textwidth}{!}{
\begin{tabular}{|c|c|l|}
\hline
  \begin{tabular}[c]{@{}c@{}}\textbf{Error}\\\textbf{Code}\end{tabular} &
  \textbf{\begin{tabular}[c]{@{}c@{}}Description \&\\ Acceptance Rate (\%)\end{tabular}} &
  \multicolumn{1}{c|}{\textbf{Example}} \\ \hline
  
  INS &
  \begin{tabular}[c]{@{}c@{}}A word is inserted.\\ 100.00\% ± 0.00\%P\end{tabular} &
  \begin{tabular}{ll}
  Original: & 고등학교 때 어긴 경험 \\
  Corrected: & 고등학교 때 \textbf{\textit{규칙을}} 어긴 경험 \\
  Translation: & Experience to break a \textbf{rule} in high school \\
  \end{tabular} \\ \hline
   
  DEL &
  \begin{tabular}[c]{@{}c@{}}A word is deleted.\\ 100.00\% ± 0.00\%P\end{tabular} &
  \begin{tabular}{ll}
  Original: & 전쟁 \textbf{\textit{끝}} 직후 장군들은 사형을 선고 받았다. \\
  Corrected: & 전쟁 직후 장군들은 사형을 선고 받았다. \\
  Translation: & After the war, the generals are sentenced to death. \\
  \end{tabular} \\ \hline
  
  WS &
  \begin{tabular}[c]{@{}c@{}}Spacing between \\ words is changed.\\100.00\% ± 0.00\%P\end{tabular} &
  \begin{tabular}{ll}
  Original: & \textbf{\textit{이옷은}} 더러워요. \\
  Corrected: & \textbf{\textit{이 옷은}} 더러워요. \\
  Translation: & This \textbf{cloth} is dirty. \\
  \end{tabular}  \\ \hline
  
  WO &
  \begin{tabular}[c]{@{}c@{}}The order of \\ words is changed.\\97.44\% ± 3.51\%P\end{tabular} &
  \begin{tabular}{ll}
  Original: & 저는 \textbf{\textit{더 한국어를}} 배우고 싶어요. \\
  Corrected: & 저는 \textbf{\textit{한국어를 더}} 배우고 싶어요. \\
  Translation: & I want to learn \textbf{Korean further}. \\
  \end{tabular} \\ \hline
  
  SPELL &
  \begin{tabular}[c]{@{}c@{}}Spelling error\\ 97.44\% ± 3.51\%P \end{tabular} &
  \begin{tabular}{ll}
  Original: & 파티에서 우리는 춤을 \textbf{\textit{쳐요.}} \\
  Corrected: & 파티에서 우리는 춤을 \textbf{\textit{춰요.}} \\
  Translation: & We \textbf{dance} at the party. \\
  \end{tabular}\\ \hline
  
  PUNCT &
  \begin{tabular}[c]{@{}c@{}}Punctuation error\\ 98.72\% ± 2.50\%P \end{tabular} &
  \begin{tabular}{ll}
  Original: & 1993년 \textbf{\textit{의}} 겨울의 일이였다. \\
  Corrected: & 1993년 \textbf{\textit{,}} 겨울의 일이였다. \\
  Translation: & It was 1993\textbf{,} a happening in winter. \\
  \end{tabular} \\ \hline
  
  SHORT &
  \begin{tabular}[c]{@{}c@{}}An edit that does not change \\ the structure of morphemes.\\ 73.08\% ± 9.84\%P \end{tabular} &
  \begin{tabular}{ll}
  Original: & 한국어는 저한테 너무 어려운 \textbf{\textit{언어이었어요.}} \\
  Corrected: & 한국어는 저한테 너무 어려운 \textbf{\textit{언어였어요.}} \\
  Translation: & Korean \textbf{Language} was too difficult to me. \\
  \end{tabular} \\ \hline
  
  VERB &
  \begin{tabular}[c]{@{}c@{}} An error on verb \\ 79.49\% ± 8.96\%P \end{tabular} &
  \begin{tabular}{ll}
  Original: & 어제 친구에게 편지를 \textbf{\textit{쌌어요.}} \\
  Corrected: & 어제 친구에게 편지를 \textbf{\textit{썼어요.}} \\
  Translation: & I \textbf{wrote} a letter to my friend yesterday. \\
  \end{tabular} \\ \hline

  ADJ &
  \begin{tabular}[c]{@{}c@{}} An error on adjective \\ 73.08\% ± 9.84\%P \end{tabular} &
  \begin{tabular}{ll}
  Original: & \textbf{\textit{진한}} 친구 \\
  Corrected: & \textbf{\textit{친한}} 친구 \\
  Translation: & A \textbf{close} friend. \\
  \end{tabular} \\ \hline

  NOUN &
  \begin{tabular}[c]{@{}c@{}} An error on noun \\ 75.64\% ± 9.53\%P\end{tabular} &
  \begin{tabular}{ll}
  Original: & 나중에 기회가 있을 때 한국에 \textbf{\textit{유학러}} 가고 싶습니다. \\
  Corrected: & 나중에 기회가 있을 때 한국에 \textbf{\textit{유학}} 가고 싶습니다. \\
  Translation: & I want to study \textbf{abroad} in Korea in the future. \\
  \end{tabular} \\ \hline

  PART &
  \begin{tabular}[c]{@{}c@{}} An error on particle \\ 97.44\% ± 3.51\%P \end{tabular} &
  \begin{tabular}{ll}
  Original: & \textbf{\textit{하와이에서}} 사는 우리 사촌 \\
  Corrected: & \textbf{\textit{하와이에}} 사는 우리 사촌 \\
  Translation: & My cousin living \textbf{in Hawaii} \\
  \end{tabular} \\ \hline

  END &
  \begin{tabular}[c]{@{}c@{}} An error on ending \\ 87.18\% ± 7.42\%P \end{tabular} &
  \begin{tabular}{ll}
  Original: & 오래 \textbf{\textit{기다려요.}} \\
  Corrected: & 오래 \textbf{\textit{기다렸어요.}} \\
  Translation: & I \textbf{waited} for a long time. \\
  \end{tabular} \\ \hline

  MOD &
  \begin{tabular}[c]{@{}c@{}} An error on modifier \\ 89.74\% ± 6.73\%P  \end{tabular} &
  \begin{tabular}{ll}
  Original: & 점심이 \textbf{\textit{나무}} 작은 나머지 배고팠어요. \\
  Corrected: & 점심이 \textbf{\textit{너무}} 작은 나머지 배고팠어요. \\
  Translation: & I was hungry because I had \textbf{such} a small launch. \\
  \end{tabular} \\ \hline

  CONJ &
  \begin{tabular}[c]{@{}c@{}} An error on conjugation \\ 43.59\% ± 11.00\%P  \end{tabular} &
  \begin{tabular}{ll}
  Original: & 오늘은 머리를 \textbf{\textit{잘라에}} 갔다. \\
  Corrected: & 오늘은 머리를 \textbf{\textit{자르러}} 갔다. \\
  Translation: & I went to a barber to get my hair \textbf{cut} today. \\
  \end{tabular} \\ \hline
\end{tabular}}
\caption{Full category of error types used in KAGAS. Middle column shows acceptance rates by each error type on human evaluation along with explanations. Rightmost column shows examples of each error types. Others are classified as UNK.}
\label{tabs:kagas_errortypes_with_example}
\end{table*}
We propose Korean Automatic Grammatical error Annotation System (KAGAS) that automatically aligns edits and annotate error types on parallel corpora that overcomes many disadvantages of hand-written annotations by human (Appendix \ref{appendix:benefits_of_kagas}). Figure \ref{fig:kagas_examples} shows the overview of KAGAS. As the scope of the system is to extract edits and annotate an error type to each edit, our system assumes the given corrected sentence is grammatically correct. Then, our system takes a pair of the original sentence and the corrected sentence as input and output aligned edits with error types. We further extend the usage of KAGAS to analyze the generated text of our baseline models by each error type in Table \ref{tabs:detail_eval_by_errortypes} at Section \ref{sec:detailed_analysis}. In this section, we describe in detail about the construction and contributions of KAGAS, with human evaluation results.
\subsection{Automatic Error Annotation for other languages}
\label{related_work:Adaptations of ERRANT onto other languages}
Creating a sufficient amount of human-annotated dataset for GEC on other languages is not trivial. To navigate this problem, there were attempts to adapt ERRANT \cite{bryant-etal-2017-automatic} onto languages other than English for error type annotation, such as on Czech \cite{naplava-etal-2022-czech}, Hindi \cite{sonawane-etal-2020-generating}, Russian \cite{katinskaia-etal-2022-semi}, German \cite{boyd-2018-using}, and Arabic \cite{belkebir-habash-2021-automatic}, but no existing work has previously extended ERRANT onto Korean. When extending ERRANT onto other languages, necessary changes about the error types were made such as discarding unmatched error types by ERRANT and adding language-specific error types.\footnote{VERB:SVA was discarded and DIACR was added for Czech \cite{naplava-etal-2022-czech} In a similar way, KAGAS made necessary changes on error types regarding the linguistic feature of Korean, such as discarding VERB:SVA and adding WS.}.

\subsection{Alignment Strategy}
Before classifying error types, we need to find \textit{where} the edits are from parallel text. We first conduct sentence-level alignment to define a "single edit". We use Damerau-Levenshtein distance \cite{felice-etal-2016-automatic} by the edit extraction repository\footnote{\url{https://github.com/chrisjbryant/edit-extraction}} to get edit pairs. Note that we apply different alignment strategy from ERRANT on the scope of a "single" edit. We use Korean-specific linguistic cost,\footnote{Kkma POS tagger, Korean lemmatizer} so that word pairs with lower POS cost and lower lemma cost are more likely to be aligned together. Also, we use custom merging rules to merge single word-level edits into WO and WS. Therefore, the number of total edits and average token length on edits, and the output $M^2$ file made from KAGAS differs from that of ERRANT, since an $M^2$ file consists of edit alignment and error type (Fig. \ref{fig:kagas_examples}). This would result in different $M^2$ scores when applied to model output evaluation.

\subsection{Error types for Korean}
We describe how we consider the unique linguistic characteristics of Korean (Appendix \ref{appendix:intro_kor}), and define 14 error types (Table
\ref{tabs:kagas_errortypes_with_example}).
\paragraph{Classifying error types in morpheme-level}
As Korean is an agglutinative language, difference between original and corrected word is naturally defined in morpheme-level. For example, 학교에 ('to school') in Table \ref{fig:kagas_examples} is divided into two parts,  학교('school') + 에('to'), based on its roles in a word. If this word is corrected to 집에('to home'), we should treat this edit as NOUN (학교 -> 집), and if this word is corrected to 학교에서('at school'), we should classify this edit as PART (Particle, since -서 is added). We need to break down original and corrected word-level edits into morphemes and only look at \textit{morpheme-level differences} between the two when classifying error types. When conducting morpheme-level alignment, we utilize morpheme-level Levenshtein distance for Korean.\footnote{\url{https://github.com/lovit/soynlp/blob/master/soynlp/hangle/_distance.py}} Also, the POS tags of Korean are based on \textit{morphemes} and not \textit{words}, meaning that there can be multiple POS tags for one word. Apart from POS tagging, KAGAS also considers the composition of edits (e.g. SHORT). Please refer to Appendix \ref{appendix:morpheme_alignment_examples} for detailed examples.

\paragraph{No PREP, but PART}
In Korean, morpheme (not word) align with the meaning. Therefore, "학교 에"(To school)->"학교에"(To-school) is WS, 
and "혁고에"(To-sceol)->"학교에"(To-school) is SPELL, which is different.
In similar vein, There is no PREP (Positioning. "to" before "school") in Korean. They rather view them as postpositional particle (Positioning "-에" after "학교")
\paragraph{On the motivation of selecting 14 Error Types}
According to previous work that categorizes Korean grammatical error by frequency \cite{shin2007corpus}, Korean error types are divided by (1) Sound, (2) Format, (3) Spacing, and (4) The rest, meaning that orthographical errors were highly frequent in Korean error types. Therefore, we designed error types to focus on capturing frequent orthographical errors such as WS, SPELL, along with syntax and morphological errors such as WO and SHORT.
There are 9 most important categories of POS for Korean (noun, pronoun, numeral, verb, adjective, postposition, pre-noun, adverbs, interjection), and a single word is divided into substantives (mostly by nouns) and inflectional words.\footnote{\url{https://www.dickgrune.com/NatLang/Korean/Lee_Chul_Young,Korean_Grammar_Textbook,indexed.pdf}} Most inflectional words are irregular and prone to change in format, and detecting those are also important. Therefore, we added 6 error types that can cover the 9 types of POS for Korean except for the numeral part\footnote{Due to lack of available data for evaluation, they are left as unclassified.} 
 (noun\&pronoun to NOUN, verb to VERB, adjective to ADJ, postposition to PART, pre-noun\&adverb to MOD, interjection to PUNCT) and 2 error types for inflectional words (CONJ, END (=suffix)), which can be classified by the POS tagger. The result of 14 error types, motivated by both Korean linguistic characteristic information in terms of linguistic typology and orthographical guidelines, contain all the crucial, frequent error types.
\begin{table*}
\centering
\small
\begin{tabular}{@{}cccccc@{}}
\toprule
\multirow{2}{*}{Dataset} & \multirow{2}{*}{Coverage}  & \multicolumn{4}{c}{Overall acceptance rate} \\
\cmidrule(r{0.125em}){3-6}
                         &                            & Total & Evaluator 1 & Evaluator 2 & Evaluator 3 \\ \midrule
%Mother Tongue            &                            & 94.24   & 93.33       & 94.82.      & 94.58   \\
%Second Learner           &                            & 87.20   & 84.52       & 87.39       & 89.68   \\ \midrule
\klearner{}             & 81.56\%  & 87.34\% ± 5.49\%P  & 84.32\% ± 10.70\%P  & 87.81\% ± 8.26\%P  & 89.88\% ± 7.44\%P  \\
\knative{}              & 90.92\%  & 93.93\% ± 2.18\%P  & 92.92\% ± 4.14\%P  & 93.99\% ± 2.82\%P  & 94.87\% ± 3.60\%P  \\
\klang{}               & 82.52\%  & 87.06\% ± 4.67\%P  & 84.72\% ± 8.88\%P  & 86.98\% ± 7.02\%P  & 89.48\% ± 6.97\%P  \\ \bottomrule
\end{tabular}
\caption{The coverage and the overall acceptance rate of KAGAS, which is a weighted sum of individual acceptance rates by error types on real dataset distributions.}
\label{tab:human-eval-result}
\end{table*}
\begin{table}
\tiny
\centering
\renewcommand{\arraystretch}{1}
\setlength{\tabcolsep}{3pt}
\resizebox{0.4\textwidth}{!}{
\begin{tabular}{@{}cc|c|c@{}}
\toprule
\multicolumn{2}{c|}{Correlation (kappa) scores}         & Value  & Reliability \\ \midrule
\multicolumn{2}{c|}{Fleiss'}                            & \textbf{0.4386} & moderate       \\ 
\multicolumn{2}{c|}{Krippendorff'}                      & \textbf{0.4392} & moderate       \\ \midrule
\multicolumn{1}{c|}{\multirow{4}{*}{\begin{tabular}[c]{@{}c@{}}Cohen's\\ (pairwise)\end{tabular}}} & Ann.1\&2 & 0.5976 & moderate       \\
\multicolumn{1}{c|}{}                         & Ann.1\&3 & 0.4426 & moderate       \\
\multicolumn{1}{c|}{}                         & Ann.2\&3 & 0.3566 & fair           \\ 
\multicolumn{1}{c|}{}                         & Average & \textbf{0.4656} & moderate        \\ \bottomrule
\end{tabular}}
\caption{Inter-annotator agreement on 3 different correlation metrics between 3 annotators. Note that Cohen suggested the kappa result be interpreted as follows: <0 indicating no agreement, <.2 none to slight,  <.4 as fair, <.6 as moderate, <.8 as substantial, and <1 as perfect agreement. The result suggest moderate agreement.}
\label{tab:inter_annotator}
\end{table}
\paragraph{About INS/DEL edits.} Since Korean is a discourse-oriented language, one can omit the subject or the object in a sentence depending on the previous context. These cases are classified as INS/DEL edits, which are grammatically correct. There are also cases of INS/DEL that edits unnecessary modifiers, which is also non-grammatical edits but rather variations to sentences. Previous works that applied ERRANT onto other languages also discard INS/DEL edits or treat them in a similar manner to our work. Works on Hindi \cite{sonawane-etal-2020-generating} and Russian \cite{katinskaia-etal-2022-semi} only classifies R: (Replacement). For Arabic \cite{belkebir-habash-2021-automatic}, Insertion and Deletion are not classified further other than token-level and word-level INS/DEL. We believe that some INS/DEL edits contain meaningful grammatical errors. However, following previous reasons and given our situation that we unfortunately don't have enough resources to conduct human evaluation of the subgroups of INS/DEL, we believe that not dividing INS/DEL any further would have more gains than losses regarding the reliability of KAGAS. DEL/INS examples are at Appendix \ref{appendix:ins_del}, and more details about selecting the granularity of error types are at Appendix Section \ref{appendix:granularity}.

\paragraph{Priority between Error Types}
Due to the nature of Korean language, multiple error types can be classified for a single edit. However, we decided to output a single representative error type for each edit (Appendix \ref{appendix:why_not_multiple_errortypes}) by defining the priority between them in order to make a deterministic, reliable system with clear evidence(Appendix \ref{appendix:kagas_error_type_priorities}).
Detailed steps are as follows: (1) We first classify edits that won’t overlap with one another (INS/DEL/WS/WO/PUNCT) according to the current error type definition. (2) After that, we prioritized classifying frequent formal and orthographical errors such as SPELL and SHORT than the rest, since those errors are highly frequent in Korean Grammar. (3) When there are single POS types for an edit, we return the error types according to the POS. (4) When there were multiple POS types per an edit, we first check whether the edit was CONJ (a combination of VERB+ENDING or ADJECTIVE+ENDING). Others are left as unclassified. We detect INS and DEL directly by the outputs of sentence-level alignment. We merge the edits for WS and WO based on the syntactical appearance of the edits. For SPELL, we use a Korean spellchecker dictionary.\footnote{\url{https://github.com/spellcheck-ko/hunspell-dict-ko}} We utilize the Korean POS tagger (Appendix \ref{appendix:kkma_pos_tagger}) to classify other POS-related error types.

\subsection{Evaluation of the Annotation System}
We evaluate our system by 3 Korean GEC experts majoring in Korean linguistics (Table \ref{tab:human-eval-result}).
First, we evaluate the acceptance rate of each error type by randomly sampling 26 parallel sentences with a single edit for each error type from our datasets. One half (13 sentences) is written by native Korean speakers, and the other is written by KFL learners.
In total, there are 364 parallel sentences in a random order. Each evaluator evaluated ``good'' or ``bad'' for each parallel sentences.
The acceptance rate is the rate of ``good'' responses out of ``good'' and ``bad'' responses\footnote{We measure the 95\% confidence interval by each error types with n=26(\#samples per error types) * 3(\#evaluators) for Table \ref{tabs:kagas_errortypes_with_example}, and report the weighted sum of those on Table \ref{tab:human-eval-result}.}.
The overall acceptance rate is the sum of the acceptance rate of each error type weighted by the proportion of the error type in each dataset.
Therefore, it depends on the distribution of error types in a dataset. By looking at the overall acceptance rate (Table \ref{tab:human-eval-result}), we can estimate that about 90\% of the classified edits are evaluated as good for KAGAS on real dataset distributions.
The coverage in a dataset is the rate of edits which is not identified as UNK (Unclassified). At Table \ref{tab:inter_annotator}, we can see that the inter-annotator agreements are moderate, meaning that the evaluation results are consistent between annotators to be reliable enough. It is also meaningful to note that, KAGAS has a very high human acceptance rate (>96.15\%) for frequently observed error types on our dataset, such as WO, SPELL, PUNCT, and PART (PARTICLE). High acceptance rate for PART is especially meaningful since PART plays an important role in representing grammatical case (-격) in Korean. Detailed analysis including the evaluation interface is at Appendix \ref{appendix:analysis_by_user_evaluation}.

\subsection{Contributions of KAGAS}
\label{sec:about_korean}
To summarize, KAGAS is different from previous work such as (1) integration of morpheme-level POS tags, (2) using morpheme-level alignment strategy, and (3) Defining Korean-specific 14 error types. Following these reasons, we believe that KAGAS capture a more diverse and accurate set of Korean error types than a simple adaptation from automatic error type systems such as \citet{sercl-choshen} or ERRANT.
\section{Experiments}
\begin{table*}[t]
\centering
\setlength\tabcolsep{4pt}
\begin{threeparttable}
\resizebox{\textwidth}{!}{
\begin{tabular}{c|c|c|c|c|c|c|c|c|c|c|c|c|c|c|c|c|c}
\toprule
 &
  \multicolumn{4}{c|}{\klearner} &
  \multicolumn{4}{c|}{\knative} &
  \multicolumn{4}{c|}{\klang} &
  \multicolumn{4}{c|}{\kunion} &
  \multirow{3}{*}{\begin{tabular}[c]{@{}c@{}}Gen.\\ time\end{tabular}}
  \\ \cline{2-17} &
  \multirow{2}{*}{GLEU} &
  \multicolumn{3}{c|}{$M^2$} &
  \multirow{2}{*}{GLEU} &
  \multicolumn{3}{c|}{$M^2$} &
  \multirow{2}{*}{GLEU} &
  \multicolumn{3}{c|}{$M^2$} &
  \multirow{2}{*}{GLEU} &
  \multicolumn{3}{c|}{$M^2$} &
  \\ \cline{3-5} \cline{7-9} \cline{11-13} \cline{15-17}
  &
   &
  Pre. &
  Rec. &
  $F_{0.5}$ &
   &
  Pre. &
  Rec. &
  $F_{0.5}$ &
   &
  Pre. &
  Rec. &
  $F_{0.5}$ &
   &
  Pre. &
  Rec. &
  $F_{0.5}$ &
   \\ \hline
\begin{tabular}[c]{@{}c@{}}Self-Scores\\[5pt] \end{tabular} &
  25.54 &
  1 &
  0 &
  0 &
  25.71 &
  1 &
  0 &
  0 &
  20.01 &
  1 &
  0 &
  0 &
  21.66 &
  1 &
  0 &
  0 &\\ 
\begin{tabular}[c]{@{}c@{}}Hanspell \\[5pt] \end{tabular} &
  30.36 &
  29.45 &
  5.33 &
  15.46 &
  57.08 &
  81.93 &
  47.36 &
  71.50 &
  22.94 &
  29.18 &
  8.74 &
  19.88 &
  28.82 &
  37.34 &
  11.58 &
  25.85 &
  189.69 \\ 
\begin{tabular}[c]{@{}c@{}}KoBART \\[5pt] \end{tabular} &
  \textbf{45.06} &
  43.35 &
  \textbf{24.54} &
  37.58 &
  \textbf{67.24} &
  75.34 &
  \textbf{55.95} &
  70.45 &
  28.48 &
  \textbf{37.56} &
  11.62 &
  25.93 &
  \textbf{33.70} &
  \textbf{44.75} &
  \textbf{14.64} &
  \textbf{31.70} &
  38.25 \\ 
\begin{tabular}[c]{@{}c@{}}KoBART +\\\kunion{}\end{tabular} &
  42.66 &
  \textbf{53.51} &
  21.18 &
  \textbf{41.00} &
  59.71 &
  \textbf{85.47} &
  47.38 &
  \textbf{73.63} &
  \textbf{28.65} &
  37.46 &
  \textbf{12.00} &
  \textbf{26.78} &
  &
  &
  &
  & 37.51
  \\ 
  \bottomrule
 %  \begin{tabular}[c]{@{}c@{}}KoBART +\\ denoising\end{tabular} &
 %  48.66 &
 %  46.11 &
 %  29.09 &
 %  41.28 &
 %  119 &
 %  71.33 &
 %  77.57 &
 %  60.62 &
 %  73.46 &
 % 69 &
 %  29.78 &
 %  38.46 &
 %  14.20 &
 %  28.59 &
 % 422 &
 %  35.66 &
 %  44.96 &
 %  17.47 &
 %  34.17 &
 % 621 \\ \hline
\end{tabular}}
\end{threeparttable}
\caption{Full experiment statistics on the test set. KoBART outputs are averaged from outputs of 3 different seeds. Generation time (sentence/second) is measured by dividing the total number of sentences by total amount of generation time taken. Accuracies on the test set is reported with model checkpoints that has the highest GLEU validation accuracy.}
\label{tabs:experiments_overall}
\end{table*}
We conduct experiments to build an effective model to encourage future research on Korean GEC models.
We report test accuracy using the model with the best validation GLEU score. 
Detailed experiment settings to reproduce our results appear in Appendix \ref{appendix:exp-details}.

\subsection{Evaluation Metrics}
\label{sec:experiments_evaluation}
We evaluate our model using $M^2$ scorer and GLEU (Section \ref{related_work:evaluation}).
Note that we can obtain $M^2$ scores as well as GLEU scores by making an $M^2$ file by KAGAS. We also report self-scores (self-GLEU and self-$M^2$, obtained by treating original text as system outputs to each evaluation system) to compare the characteristics of the dataset itself. Higher self-scores would mean that corrected text is similar to the original text.

\subsection{Baseline GEC system, Hanspell}
%Since we are proposing a new dataset and existing previous works lack instructions for replication, o
Our primary aim is to build a first strong baseline model on Korean GEC. Therefore, we compare our methods with a commercial, well-known Korean GEC statistical system called Hanspell.\footnote{\url{https://github.com/9beach/hanspell}} (Note that Hanspell is a completely different system from \textit{Hunspell}.\footnote{\url{https://github.com/hunspell/hunspell}} a spellchecker) It is developed by the Pusan University since 1992 and it widely used in Korea since it is free and easily accessible through the web.\footnote{\url{https://speller.cs.pusan.ac.kr/}}

\subsection{Dataset split}
We split datasets by the train\_test\_val\_split function from sklearn.\footnote{\url{https://scikit-learn.org/stable/}} The train, test, valid ratio is set to 70\%, 15\%, and 15\% on seed 0. Then, we aggregate all three datasets to make \textit{\kunion{}}.
\footnote{For fair evaluation, all 3 datasets are evenly distributed for each train, test, and valid split of \kunion{}.} 

\subsection{Model Training}
We use the HuggingFace\footnote{\url{https://huggingface.co/transformers}} implementation of BART by loading the weights from the pre-trained Korean BART (KoBART).
We train models with multiple scenarios: (1) fine-tuning KoBART with 3 individual datasets, and (2) fine-tuning with \kunion{} and additionally fine-tuning on top of it with 3 individual data. We run each model with three different seeds and report the average score. For (1), we use a learning rate of 3e-5 for 10 epochs with a batch size of 64 for all datasets on a TESLA V100 13GB GPU. Other hyperparameters are the same as KoBART configurations. For (2), we use a learning rate of 1e-5.

\subsection{Tokenization} 
We utilize the character BPE \cite{sennrich-etal-2016-neural} tokenizer from HuggingFace tokenizers library, as KoBART used. Due to the limitations of the tokenizer, the encoded then decoded version of the original raw text automatically removes spaces between a word and punctuation (Appendix \ref{appendix:tokenizer_issue}). Therefore, naive evaluation of the generated output (decoded by the tokenizer) with the $M^2$ file made by raw text output is not aligned well, resulting in bad accuracy. Since we thought measuring the performance of the model has higher priority than measuring the performance of the tokenizer, we use the decoded version of text to train and make $M^2$ files for evaluation.

\section{Results and Discussion}
% \usepackage{booktabs}
\begin{table*}[t]
\centering
\setlength{\tabcolsep}{5pt}
\resizebox{\textwidth}{!}{
\begin{tabular}{@{}c|c|c|c|ccccccccccccccc@{}}
\toprule
 &
   &
  FULL&
  STD&
  INS&
  DEL&
  WS&
  WO&
  SPELL&
  PUN.&
  SHO.&
  VERB&
  ADJ&
  NOUN&
  PART&
  END&
  MOD&
  CONJ&
  UNK\\ \midrule
\multicolumn{1}{c|}{GLEU} &
  Hanspell &
  28.82&
  9.19&
  \textbf{21.89}&
  13.57&
  \textbf{46.91}&
  7.51&
  31.73&
  16.81&
  31.06&
  19.94&
  18.31&
  19.67&
  17.14&
  19.17&
  17.41&
  19.79&
  \textbf{21.15}\\
\multicolumn{1}{c|}{\textbf{> 20}} &
  KoBART &
  33.70&
  6.02&
  \textbf{23.67}&
  18.22&
  \textbf{34.83}&
  13.76&
  \textbf{36.10}&
  \textbf{20.29}&
  \textbf{33.42}&
  \textbf{27.96}&
  \textbf{29.27}&
  \textbf{25.09}&
  \textbf{27.15}&
  \textbf{27.72}&
  \textbf{23.68}&
  \textbf{25.95}&
  \textbf{26.44}\\ \midrule
\multicolumn{1}{c|}{Prec.} &
  Hanspell &
  37.34&
  12.22&
  37.49&
  33.51&
  \textbf{71.98}&
  25.00&
  \textbf{48.21}&
  30.61&
  \textbf{46.70}&
  30.32&
  28.42&
  29.94&
  24.37&
  27.12&
  31.41&
  32.10&
  34.07\\
\multicolumn{1}{c|}{\textbf{> 40}} &
  KoBART &
  44.75&
  6.78&
  \textbf{40.58}&
  \textbf{45.64}&
  \textbf{63.26}&
  37.89&
  \textbf{52.16}&
  \textbf{42.35}&
  \textbf{58.52}&
  \textbf{46.45}&
  \textbf{49.16}&
  \textbf{43.76}&
  \textbf{49.35}&
  \textbf{47.51}&
  \textbf{46.67}&
  \textbf{41.30}&
  \textbf{42.69}\\ \midrule
\multicolumn{1}{c|}{Recall} &
  Hanspell &
  11.58&
  9.00&
  \textbf{10.10}&
  7.40&
  \textbf{39.30}&
  5.26&
  \textbf{21.25}&
  6.14&
  \textbf{15.48}&
  6.62&
  6.12&
  8.14&
  5.01&
  5.64&
  7.08&
  8.60&
  9.69\\
\multicolumn{1}{c|}{\textbf{> 10}} &
  KoBART &
  14.64&
  4.09&
  9.47&
  \textbf{12.05}&
  \textbf{21.48}&
  9.04&
  \textbf{22.54}&
  \textbf{10.06}&
  \textbf{19.61}&
  \textbf{13.29}&
  \textbf{13.92}&
  \textbf{12.46}&
  \textbf{14.30}&
  \textbf{13.23}&
  \textbf{12.70}&
  \textbf{12.34}&
  \textbf{12.68}\\ \midrule
\multicolumn{1}{c|}{$F_{0.5}$} &
  Hanspell &
  25.85&
  12.58&
  24.31&
  19.64&
  \textbf{61.72}&
  14.29&
  \textbf{38.46}&
  17.03&
  \textbf{33.28}&
  17.67&
  16.43&
  19.50&
  13.75&
  15.40&
  18.61&
  20.76&
  22.66\\
\multicolumn{1}{c|}{\textbf{> 30}} &
  KoBART &
  31.70&
  6.46&
  24.47&
  29.28&
  \textbf{45.48}&
  23.11&
  \textbf{41.30}&
  25.78&
  \textbf{41.84}&
  \textbf{30.98}&
  \textbf{32.61}&
  29.11&
  \textbf{33.10}&
  \textbf{31.28}&
  \textbf{30.38}&
  28.09&
  28.97\\ \bottomrule
\end{tabular}}
\caption{GLEU and \mm{} scores on the generation output on the test set of Hanspell and KoBART on Kor-Union. The scores are divided by all 14 + UNK error types. For convenience, scores higher than a certain threshold are highlighted. PUN. and SHO. is for PUNCT and SHORT, respectively. The standard deviation(STD) of Hanspell is higher than that of KoBART, meaning that \textbf{scores by KoBART are evenly distributed for all error types}, while scores by Hanspell are biased toward WS and SPELL.}
\label{tabs:detail_eval_by_errortypes}
\end{table*}
\paragraph{Effectiveness of Neural models}
As we can see in Table \ref{tabs:experiments_overall}, the model trained with our dataset outperform the current commercial GEC system (Hanspell) on all datasets. It is notable in that Hanspell is currently known as the best performing system open-source system for correcting erroneous Korean sentences. The result implies that our dataset helps to build a better GEC system, and our that our model can serve as a reasonable baseline that shows the effectiveness of neural models against previous rule-based systems on GEC. Moreover, the generation speed of our neural models (KoBART) is about five times faster than Hanspell, showing the efficiency as well as performance.\footnote{Note that we used the \textit{hanspell-cli github wrapper} and not the \textit{browser} to compare inference speed, for fair comparison. \url{https://github.com/9beach/hanspell}}
\paragraph{Analysis by Error Types}
\label{sec:detailed_analysis}
Here, we demonstrate the usefulness of KAGAS, which enables us to conduct a detailed post-analysis of model output by measuring model performance on individual error types. %KAGAS can be served as a valuable resource to evaluate the performance of different models qualitatively from various perspectives. 
Table \ref{tabs:detail_eval_by_errortypes} shows score distributions on individual error types for Hanspell and KoBART on Kor-Union. (Full scores are at Appendix \ref{tabs:errortype_scores_on_everything}). Compared with Hanspell, KoBART trained with our dataset generally perform better regardless of error types. In contrast, Hanspell's performance is very biased towards SPELL and WS. 
\paragraph{\knative{}}
Note that the performance of \knative{} is much higher than the other datasets. The error type distribution (Figure \ref{fig:errortype_distribution}) for \knative{} aligns with \citet{shin2015} that more than half of the dataset is on WS for native Korean speakers, which is different from learner datasets which has a more diverse set of error types. Therefore, it is easier for the model to train on \knative{} than on other datasets.

\section{Conclusion}
In this work, we (1) construct three parallel datasets of grammatically incorrect and corrected Korean sentence pairs for training Korean GEC systems: \klang{}, \knative{}, and \klearner{}. Our datasets are complementary representing grammatical errors that generated by both native Korean speakers and KFL learners.
(2) to train and evaluate models with these new datasets, we develop KAGAS, which considers the linguistic characteristic of Korean and automatically aligns and annotates edits between sentence pairs.
(3) We show our experimental results based on a pre-trained KoBART model with fine-tuning on our datasets and compare them with a baseline system, Hanspell.
We expect that our datasets, evaluation toolkit, and models will foster active future research on Korean GEC as well as a wide range of Korean NLP tasks. Future work includes on further refining our proposed method, KAGAS, by extending the coverage and making more accurate error type classification.
\clearpage

\section*{Limitations}
Our automatic error type system, KAGAS, has room for improvements. Although we got high human acceptance rate for the error type classification results of KAGAS, Our coverage of error types is about 80\% to 90\% (Table \ref{tab:human-eval-result}). Currently, our system rely on the Kkma POS tagger for Korean. We believe that the improvement of a POS tagger will enable KAGAS to define a more detailed error type classification with high coverage and reliability. Also, there could be other ways (or more efficient ways) to define and classify Korean grammatical edits. However, we would like KAGAS to be viewed as the first step towards the effort of making an automatic annotation tool for Korean GEC, which, though not perfect, have meaningful contributions to the field in its current form.

\textbf{Future Directions.} Currently, the 14 error types of KAGAS is focused to be as specifc as possible, while respecting both statistical characteristics of Korean language and incorporation into a reliable, deterministic system with high agreement of human evaluation. However, definment of a more richer error type classfication system derived from KAGAS such as differentiating between the typographical and phonetic errors would be an important future direction for our research, as both are defined as SPELL errors on our current system. It would require solving additional challenges of accurately disambiguating a writer's intention behind errors on a grammatical aspect. Another possible future direction would be applying data augmentation techniques on our datasets to boost the size of the training examples and obtain evaluation metric accuracy gains. 

\section*{Ethics Statement}
We have conducted an IRB for KAGAS human evaluation.\footnote{Approval number: KH2021-020}

\section*{Acknowledgement}
We would like to thank all the LKLab lab mates, and Prof. Jungyeul Park for helpful discussions. We would also like to thank Geunhoo Kim, Jaeyun Kim, and from Sumin Lim for helping the dataset collection process and making the primary version of this paper. Lastly, I would like to thank Jinwoo Kim for helping me write a better rebuttal, and all the anonymous reviewers who really helped the most to make the current version of the paper.

This work was partly supported by Institute of Information \& communications Technology Planning \& Evaluation (IITP) grant funded by the Korea government (MSIT) (No. 2022-0-00184, Development and Study of AI Technologies to Inexpensively Conform to Evolving Policy on Ethics, 70\%; No.2021-0-02068, Artificial Intelligence Innovation Hub, 30\%).

\bibliography{acl}

\begin{thebibliography}{48}
\expandafter\ifx\csname natexlab\endcsname\relax\def\natexlab#1{#1}\fi

\bibitem[{Belkebir and Habash(2021)}]{belkebir-habash-2021-automatic}
Riadh Belkebir and Nizar Habash. 2021.
\newblock \href {https://doi.org/10.18653/v1/2021.conll-1.47} {Automatic error
  type annotation for {A}rabic}.
\newblock In \emph{Proceedings of the 25th Conference on Computational Natural
  Language Learning}, pages 596--606, Online. Association for Computational
  Linguistics.

\bibitem[{Bender(2011)}]{bender2011achieving}
Emily~M Bender. 2011.
\newblock On achieving and evaluating language-independence in nlp.
\newblock \emph{Linguistic Issues in Language Technology}, 6(3):1--26.

\bibitem[{Boyd(2018)}]{boyd-2018-using}
Adriane Boyd. 2018.
\newblock \href {https://doi.org/10.18653/v1/W18-6111} {Using {W}ikipedia edits
  in low resource grammatical error correction}.
\newblock In \emph{Proceedings of the 2018 {EMNLP} Workshop W-{NUT}: The 4th
  Workshop on Noisy User-generated Text}, pages 79--84, Brussels, Belgium.
  Association for Computational Linguistics.

\bibitem[{Bryant et~al.(2019)Bryant, Felice, Andersen, and
  Briscoe}]{bryant-etal-2019-bea}
Christopher Bryant, Mariano Felice, {\O}istein~E. Andersen, and Ted Briscoe.
  2019.
\newblock \href {https://doi.org/10.18653/v1/W19-4406} {The bea-2019 shared
  task on grammatical error correction}.
\newblock In \emph{Proceedings of the Fourteenth Workshop on Innovative Use of
  NLP for Building Educational Applications}, pages 52--75, Florence, Italy.
  Association for Computational Linguistics.

\bibitem[{Bryant et~al.(2017)Bryant, Felice, and
  Briscoe}]{bryant-etal-2017-automatic}
Christopher Bryant, Mariano Felice, and Ted Briscoe. 2017.
\newblock \href {https://doi.org/10.18653/v1/P17-1074} {Automatic annotation
  and evaluation of error types for grammatical error correction}.
\newblock In \emph{Proceedings of the 55th Annual Meeting of the Association
  for Computational Linguistics (Volume 1: Long Papers)}, pages 793--805,
  Vancouver, Canada. Association for Computational Linguistics.

\bibitem[{Choshen and Abend(2018)}]{sercl-choshen}
Leshem Choshen and Omri Abend. 2018.
\newblock \href {http://arxiv.org/abs/1804.11225} {Automatic metric validation
  for grammatical error correction}.
\newblock \emph{CoRR}, abs/1804.11225.

\bibitem[{Comrie(1989)}]{comrie1989language}
Bernard Comrie. 1989.
\newblock \emph{Language universals and linguistic typology: Syntax and
  morphology}.
\newblock University of Chicago press.

\bibitem[{Dahlmeier and Ng(2012)}]{dahlmeier-ng-2012-better}
Daniel Dahlmeier and Hwee~Tou Ng. 2012.
\newblock \href {https://www.aclweb.org/anthology/N12-1067} {Better evaluation
  for grammatical error correction}.
\newblock In \emph{Proceedings of the 2012 Conference of the North {A}merican
  Chapter of the Association for Computational Linguistics: Human Language
  Technologies}, pages 568--572, Montr{\'e}al, Canada. Association for
  Computational Linguistics.

\bibitem[{Dickinson et~al.(2011)Dickinson, Israel, and
  Lee}]{dickinson-etal-2011-developing}
Markus Dickinson, Ross Israel, and Sun-Hee Lee. 2011.
\newblock \href {https://www.aclweb.org/anthology/W11-1410} {Developing
  methodology for {K}orean particle error detection}.
\newblock In \emph{Proceedings of the Sixth Workshop on Innovative Use of {NLP}
  for Building Educational Applications}, pages 81--86, Portland, Oregon.
  Association for Computational Linguistics.

\bibitem[{Felice et~al.(2016)Felice, Bryant, and
  Briscoe}]{felice-etal-2016-automatic}
Mariano Felice, Christopher Bryant, and Ted Briscoe. 2016.
\newblock \href {https://www.aclweb.org/anthology/C16-1079} {Automatic
  extraction of learner errors in {ESL} sentences using linguistically enhanced
  alignments}.
\newblock In \emph{Proceedings of {COLING} 2016, the 26th International
  Conference on Computational Linguistics: Technical Papers}, pages 825--835,
  Osaka, Japan. The COLING 2016 Organizing Committee.

\bibitem[{Grundkiewicz and Junczys-Dowmunt(2014)}]{grundkiewicz2014wiked}
Roman Grundkiewicz and Marcin Junczys-Dowmunt. 2014.
\newblock The wiked error corpus: A corpus of corrective wikipedia edits and
  its application to grammatical error correction.
\newblock In \emph{International Conference on Natural Language Processing},
  pages 478--490. Springer.

\bibitem[{Haupt et~al.(2017)Haupt, Alt, and Buttrey}]{haupt2017identifying}
Adam~Christian Haupt, Jonathan Alt, and Samuel Buttrey. 2017.
\newblock Identifying students at risk in academics: Analysis of korean
  language academic attrition at the defense language institute foreign
  language center.
\newblock \emph{Journal of Defense Analytics and Logistics}.

\bibitem[{Israel et~al.(2013)Israel, Dickinson, and
  Lee}]{israel-etal-2013-detecting}
Ross Israel, Markus Dickinson, and Sun-Hee Lee. 2013.
\newblock \href {https://www.aclweb.org/anthology/I13-1199} {Detecting and
  correcting learner {K}orean particle omission errors}.
\newblock In \emph{Proceedings of the Sixth International Joint Conference on
  Natural Language Processing}, pages 1419--1427, Nagoya, Japan. Asian
  Federation of Natural Language Processing.

\bibitem[{Katinskaia et~al.(2022)Katinskaia, Lebedeva, Hou, and
  Yangarber}]{katinskaia-etal-2022-semi}
Anisia Katinskaia, Maria Lebedeva, Jue Hou, and Roman Yangarber. 2022.
\newblock \href {https://aclanthology.org/2022.lrec-1.88} {Semi-automatically
  annotated learner corpus for {R}ussian}.
\newblock In \emph{Proceedings of the Thirteenth Language Resources and
  Evaluation Conference}, pages 832--839, Marseille, France. European Language
  Resources Association.

\bibitem[{Katsumata and Komachi(2020)}]{katsumata-komachi-2020-stronger}
Satoru Katsumata and Mamoru Komachi. 2020.
\newblock \href {https://www.aclweb.org/anthology/2020.aacl-main.83} {Stronger
  baselines for grammatical error correction using a pretrained encoder-decoder
  model}.
\newblock In \emph{Proceedings of the 1st Conference of the Asia-Pacific
  Chapter of the Association for Computational Linguistics and the 10th
  International Joint Conference on Natural Language Processing}, pages
  827--832, Suzhou, China. Association for Computational Linguistics.

\bibitem[{Kim(2020)}]{kim2020kfl}
Gyeongyeol Kim. 2020.
\newblock Foreign students' korean writing errors - studying methods of
  language.
\newblock \emph{The Journal of Language and Literature}, 82:363--389.

\bibitem[{Koyama et~al.(2020)Koyama, Kiyuna, Kobayashi, Arai, and
  Komachi}]{koyama-etal-2020-construction}
Aomi Koyama, Tomoshige Kiyuna, Kenji Kobayashi, Mio Arai, and Mamoru Komachi.
  2020.
\newblock \href {https://www.aclweb.org/anthology/2020.lrec-1.26} {Construction
  of an evaluation corpus for grammatical error correction for learners of
  {J}apanese as a second language}.
\newblock In \emph{Proceedings of The 12th Language Resources and Evaluation
  Conference}, pages 204--211, Marseille, France. European Language Resources
  Association.

\bibitem[{Lee(2018)}]{lee2018effects}
Inhye Lee. 2018.
\newblock Effects of contact with korean popular culture on kfl learners'
  motivation.
\newblock \emph{The Korean Language in America}, 22(1):25--45.

\bibitem[{Lee(2014)}]{OUNHCHR2014}
Kwankyu Lee. 2014.
\newblock \emph{Hangeul Orthography Impact Assessment}.
\newblock 11-1371028-000540-01. KCI.

\bibitem[{Lee(2020)}]{lee2020kor}
Minwoo Lee. 2020.
\newblock An analysis of korean learners' errors by proficiency: Focus on
  statistical analysis using the multinomial logistic regression model.
\newblock \emph{Journal of Korean Language Education}, 31(2):143--169.

\bibitem[{Lee et~al.(2021)Lee, Shin, Lee, and Choi}]{lee2021korean}
Myunghoon Lee, Hyeonho Shin, Dabin Lee, and Sung-Pil Choi. 2021.
\newblock Korean grammatical error correction based on transformer with copying
  mechanisms and grammatical noise implantation methods.
\newblock \emph{Sensors}.

\bibitem[{Lee et~al.(2012)Lee, Dickinson, and
  Israel}]{lee-etal-2012-developing}
Sun-Hee Lee, Markus Dickinson, and Ross Israel. 2012.
\newblock \href {https://www.aclweb.org/anthology/W12-3617} {Developing learner
  corpus annotation for {K}orean particle errors}.
\newblock In \emph{Proceedings of the Sixth Linguistic Annotation Workshop},
  pages 129--133, Jeju, Republic of Korea. Association for Computational
  Linguistics.

\bibitem[{Lee and Song(2012)}]{lee2012annotating}
Sun-Hee Lee and Jae-young Song. 2012.
\newblock Annotating particle realization and ellipsis in korean.
\newblock In \emph{Proceedings of the Sixth Linguistic Annotation Workshop},
  pages 175--183.

\bibitem[{Levenshtein(1965)}]{levenshtein1966binary}
Vladimir~I. Levenshtein. 1965.
\newblock Binary codes capable of correcting deletions, insertions, and
  reversals.
\newblock \emph{Soviet physics. Doklady}, 10:707--710.

\bibitem[{Lewis et~al.(2019)Lewis, Liu, Goyal, Ghazvininejad, Mohamed, Levy,
  Stoyanov, and Zettlemoyer}]{lewis2019bart}
Mike Lewis, Yinhan Liu, Naman Goyal, Marjan Ghazvininejad, Abdelrahman Mohamed,
  Omer Levy, Ves Stoyanov, and Luke Zettlemoyer. 2019.
\newblock Bart: Denoising sequence-to-sequence pre-training for natural
  language generation, translation, and comprehension.
\newblock \emph{arXiv preprint arXiv:1910.13461}.

\bibitem[{Lewis et~al.(2020)Lewis, Liu, Goyal, Ghazvininejad, Mohamed, Levy,
  Stoyanov, and Zettlemoyer}]{lewis-etal-2020-bart}
Mike Lewis, Yinhan Liu, Naman Goyal, Marjan Ghazvininejad, Abdelrahman Mohamed,
  Omer Levy, Veselin Stoyanov, and Luke Zettlemoyer. 2020.
\newblock \href {https://doi.org/10.18653/v1/2020.acl-main.703} {{BART}:
  Denoising sequence-to-sequence pre-training for natural language generation,
  translation, and comprehension}.
\newblock In \emph{Proceedings of the 58th Annual Meeting of the Association
  for Computational Linguistics}, pages 7871--7880, Online. Association for
  Computational Linguistics.

\bibitem[{Li et~al.(2018)Li, Zhou, Bao, Liu, Xu, and Li}]{li-etal-2018-hybrid}
Chen Li, Junpei Zhou, Zuyi Bao, Hengyou Liu, Guangwei Xu, and Linlin Li. 2018.
\newblock \href {https://doi.org/10.18653/v1/W18-3708} {A hybrid system for
  {C}hinese grammatical error diagnosis and correction}.
\newblock In \emph{Proceedings of the 5th Workshop on Natural Language
  Processing Techniques for Educational Applications}, pages 60--69, Melbourne,
  Australia. Association for Computational Linguistics.

\bibitem[{Lichtarge et~al.(2019)Lichtarge, Alberti, Kumar, Shazeer, Parmar, and
  Tong}]{lichtarge-etal-2019-corpora}
Jared Lichtarge, Chris Alberti, Shankar Kumar, Noam Shazeer, Niki Parmar, and
  Simon Tong. 2019.
\newblock \href {https://doi.org/10.18653/v1/N19-1333} {Corpora generation for
  grammatical error correction}.
\newblock In \emph{Proceedings of the 2019 Conference of the North {A}merican
  Chapter of the Association for Computational Linguistics: Human Language
  Technologies, Volume 1 (Long and Short Papers)}, pages 3291--3301,
  Minneapolis, Minnesota. Association for Computational Linguistics.

\bibitem[{Min et~al.(2020)Min, Jung, Jung, Yang, Cho, and
  Kim}]{min2020grammatical}
Jin~Hong Min, Seong~Jun Jung, Se~Hee Jung, Seongmin Yang, Jun~Sang Cho, and
  Sung~Hwan Kim. 2020.
\newblock Grammatical error correction models for korean language via
  pre-trained denoising.
\newblock \emph{Quantitative Bio-Science}, pages 17--24.

\bibitem[{Mizumoto et~al.(2011)Mizumoto, Komachi, Nagata, and
  Matsumoto}]{mizumoto2011mining}
Tomoya Mizumoto, Mamoru Komachi, Masaaki Nagata, and Yuji Matsumoto. 2011.
\newblock Mining revision log of language learning sns for automated japanese
  error correction of second language learners.
\newblock In \emph{Proceedings of 5th International Joint Conference on Natural
  Language Processing}, pages 147--155.

\bibitem[{Nam et~al.(2018)Nam, Yoo, and Hong}]{Nam2018}
Yunju Nam, Jewook Yoo, and Upyong Hong. 2018.
\newblock The influence of constituents' semantic properties on the word order
  preference in korean sentence production.
\newblock \emph{Language and Information}, 22(1).

\bibitem[{N{\'a}plava et~al.(2022)N{\'a}plava, Straka, Strakov{\'a}, and
  Rosen}]{naplava-etal-2022-czech}
Jakub N{\'a}plava, Milan Straka, Jana Strakov{\'a}, and Alexandr Rosen. 2022.
\newblock \href {https://doi.org/10.1162/tacl_a_00470} {{C}zech grammar error
  correction with a large and diverse corpus}.
\newblock \emph{Transactions of the Association for Computational Linguistics},
  10:452--467.

\bibitem[{Napoles et~al.(2015)Napoles, Sakaguchi, Post, and
  Tetreault}]{napoles-etal-2015-ground}
Courtney Napoles, Keisuke Sakaguchi, Matt Post, and Joel Tetreault. 2015.
\newblock \href {https://doi.org/10.3115/v1/P15-2097} {Ground truth for
  grammatical error correction metrics}.
\newblock In \emph{Proceedings of the 53rd Annual Meeting of the Association
  for Computational Linguistics and the 7th International Joint Conference on
  Natural Language Processing (Volume 2: Short Papers)}, pages 588--593,
  Beijing, China. Association for Computational Linguistics.

\bibitem[{Ng et~al.(2014)Ng, Wu, Briscoe, Hadiwinoto, Susanto, and
  Bryant}]{ng-etal-2014-conll}
Hwee~Tou Ng, Siew~Mei Wu, Ted Briscoe, Christian Hadiwinoto, Raymond~Hendy
  Susanto, and Christopher Bryant. 2014.
\newblock \href {https://doi.org/10.3115/v1/W14-1701} {The {C}o{NLL}-2014
  shared task on grammatical error correction}.
\newblock In \emph{Proceedings of the Eighteenth Conference on Computational
  Natural Language Learning: Shared Task}, pages 1--14, Baltimore, Maryland.
  Association for Computational Linguistics.

\bibitem[{Papineni et~al.(2002)Papineni, Roukos, Ward, and
  Zhu}]{papineni-etal-2002-bleu}
Kishore Papineni, Salim Roukos, Todd Ward, and Wei-Jing Zhu. 2002.
\newblock \href {https://doi.org/10.3115/1073083.1073135} {{B}leu: a method for
  automatic evaluation of machine translation}.
\newblock In \emph{Proceedings of the 40th Annual Meeting of the Association
  for Computational Linguistics}, pages 311--318, Philadelphia, Pennsylvania,
  USA. Association for Computational Linguistics.

\bibitem[{Park et~al.(2020)Park, Park, and Lim}]{park2020self}
Chanjun Park, Sungjin Park, and Heuiseok Lim. 2020.
\newblock Self-supervised korean spelling correction via denoising transformer.
\newblock \emph{7th International Conference on Information, System, and
  Convergence Applications}.

\bibitem[{Rao et~al.(2018)Rao, Gong, Zhang, and Xun}]{rao-etal-2018-overview}
Gaoqi Rao, Qi~Gong, Baolin Zhang, and Endong Xun. 2018.
\newblock \href {https://doi.org/10.18653/v1/W18-3706} {Overview of
  {NLPTEA}-2018 share task {C}hinese grammatical error diagnosis}.
\newblock In \emph{Proceedings of the 5th Workshop on Natural Language
  Processing Techniques for Educational Applications}, pages 42--51, Melbourne,
  Australia. Association for Computational Linguistics.

\bibitem[{Rozovskaya and Roth(2019)}]{rozovskaya-roth-2019-grammar}
Alla Rozovskaya and Dan Roth. 2019.
\newblock \href {https://doi.org/10.1162/tacl_a_00251} {Grammar error
  correction in morphologically rich languages: The case of {R}ussian}.
\newblock \emph{Transactions of the Association for Computational Linguistics},
  7:1--17.

\bibitem[{Sennrich et~al.(2016)Sennrich, Haddow, and
  Birch}]{sennrich-etal-2016-neural}
Rico Sennrich, Barry Haddow, and Alexandra Birch. 2016.
\newblock \href {https://doi.org/10.18653/v1/P16-1162} {Neural machine
  translation of rare words with subword units}.
\newblock In \emph{Proceedings of the 54th Annual Meeting of the Association
  for Computational Linguistics (Volume 1: Long Papers)}, pages 1715--1725,
  Berlin, Germany. Association for Computational Linguistics.

\bibitem[{Shin et~al.(2015)Shin, Kim, and Lee}]{shin2015}
Hocheol Shin, Buyeon Kim, and Kyubum Lee. 2015.
\newblock A study on the hangeul orthography error status.
\newblock \emph{Grammar Education}, 23:63--94.

\bibitem[{Shin(2007)}]{shin2007corpus}
Seoin Shin. 2007.
\newblock Corpus-based study of word order variations in korean.
\newblock In \emph{Proceedings of the Corpus Linguistics Conference (CL2007)},
  volume 2730. Citeseer.

\bibitem[{Sohn(2001)}]{sohn2001korean}
Ho-Min Sohn. 2001.
\newblock \emph{The Korean language}.
\newblock Cambridge University Press.

\bibitem[{Sonawane et~al.(2020)Sonawane, Vishwakarma, Srivastava, and
  Kumar~Singh}]{sonawane-etal-2020-generating}
Ankur Sonawane, Sujeet~Kumar Vishwakarma, Bhavana Srivastava, and Anil
  Kumar~Singh. 2020.
\newblock \href {https://aclanthology.org/2020.aacl-srw.24} {Generating
  inflectional errors for grammatical error correction in {H}indi}.
\newblock In \emph{Proceedings of the 1st Conference of the Asia-Pacific
  Chapter of the Association for Computational Linguistics and the 10th
  International Joint Conference on Natural Language Processing: Student
  Research Workshop}, pages 165--171, Suzhou, China. Association for
  Computational Linguistics.

\bibitem[{Song(2006)}]{song2006korean}
Jae~Jung Song. 2006.
\newblock \emph{The Korean language: Structure, use and context}.
\newblock Routledge.

\bibitem[{Tsarfaty et~al.(2010)Tsarfaty, Seddah, Goldberg, K{\"u}bler, Versley,
  Candito, Foster, Rehbein, and Tounsi}]{tsarfaty2010statistical}
Reut Tsarfaty, Djam{\'e} Seddah, Yoav Goldberg, Sandra K{\"u}bler, Yannick
  Versley, Marie Candito, Jennifer Foster, Ines Rehbein, and Lamia Tounsi.
  2010.
\newblock Statistical parsing of morphologically rich languages (spmrl) what,
  how and whither.
\newblock In \emph{Proceedings of the NAACL HLT 2010 First Workshop on
  Statistical Parsing of Morphologically-Rich Languages}, pages 1--12.

\bibitem[{Vania and Lopez(2017)}]{vania2017characters}
Clara Vania and Adam Lopez. 2017.
\newblock From characters to words to in between: Do we capture morphology?
\newblock \emph{arXiv preprint arXiv:1704.08352}.

\bibitem[{Wu et~al.(2018)Wu, Wang, Chen, and Yang}]{wu-etal-2018-cyut}
Shih-Hung Wu, Jun-Wei Wang, Liang-Pu Chen, and Ping-Che Yang. 2018.
\newblock \href {https://doi.org/10.18653/v1/W18-3729} {{CYUT}-{III} team
  {C}hinese grammatical error diagnosis system report in {NLPTEA}-2018 {CGED}
  shared task}.
\newblock In \emph{Proceedings of the 5th Workshop on Natural Language
  Processing Techniques for Educational Applications}, pages 199--202,
  Melbourne, Australia. Association for Computational Linguistics.

\bibitem[{Zhao et~al.(2019)Zhao, Wang, Shen, Jia, and
  Liu}]{zhao-etal-2019-improving}
Wei Zhao, Liang Wang, Kewei Shen, Ruoyu Jia, and Jingming Liu. 2019.
\newblock \href {https://doi.org/10.18653/v1/N19-1014} {Improving grammatical
  error correction via pre-training a copy-augmented architecture with
  unlabeled data}.
\newblock In \emph{Proceedings of the 2019 Conference of the North {A}merican
  Chapter of the Association for Computational Linguistics: Human Language
  Technologies, Volume 1 (Long and Short Papers)}, pages 156--165, Minneapolis,
  Minnesota. Association for Computational Linguistics.

\end{thebibliography}
\bibliographystyle{acl_natbib}
\clearpage

\clearpage
\appendix

\numberwithin{table}{section}
\setcounter{page}{1}

\begin{appendix}
\onecolumn

\begin{center}
  \noindent \textbf{\Large Appendix for \\Standardizing Korean Grammatical Error Correction: \\Datasets, Evaluation, and Models}
\end{center}

%%%Data stats%%%%%%%%%%%%%%%%%%%%%%%%%%%%%%%%%%%%%%%
\section{Detailed instruction of dataset pre-processing}

\subsection{General}
In original-corrected pairs, there are cases where punctuation and words are in one word for origial-corrected edit pairs, such as: "갔어." -> "갔어!" Since we are doing a word-level alignment, it seems inappropirate to classify this whole edit as "PUNCT". Therefore, in order to correctly get error type distributions per our dataset, we process all of our dataset to add spaces between punctuations ("갔어." -> "갔어 .", "갔어!" -> "갔어 !"). After this, only punctuations can be left for alignment. Now the edit pairs from the previous example are transformed into "."->"!", which seems very appropriate as an edit that can be classified as "PUNCT".

\subsection{\knative}
\label{appendix:native_collection}
\noindent \textbf{Collecting correct sentences.}
We collect grammatically correct sentences from two sources:
\begin{enumerate}
    \item 7,481 sentences from online education materials for Korean learners published by the Center for Teaching and Learning for Korean.\footnote{\url{https://kcorpus.korean.go.kr/service/goErrorAnnotationSearch.do}}
    \item 4,182 example sentences in an electronic dictionary written by NIKL.\footnote{These sentences come from the National Institute of Korean Language which allows use of this corpus for research purposes.}
\end{enumerate}
We have granted to apply changes to the original dataset (Additionally make grammatically wrong sentences out of correct sentence) and redistribute these datasets, under the Korean Gong-Gong-Nuri-4 license.\footnote{\url{https://www.kogl.or.kr/info/license.do\#04-tab}} This license states that anyone can use \knative{} for non-commercial purposes under proper attribution of source.

\noindent \textbf{Collecting transcribed sentences.}
\begin{figure}[!h]
\centering
\includegraphics[width=\linewidth]{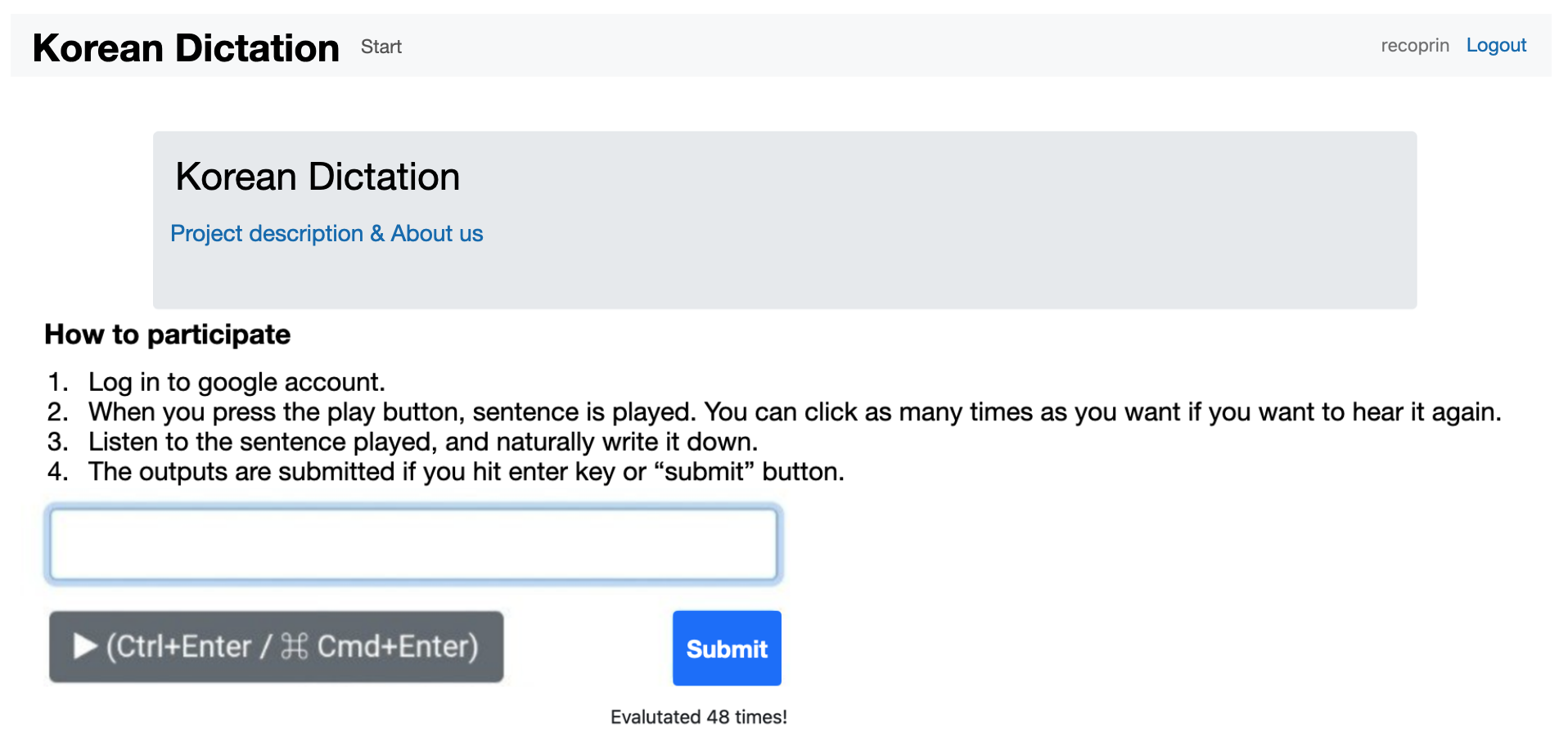}
\label{fig:nativedemo}
\caption{Demo page that we used for \knative{} dataset collection. Translated into english.}
\end{figure}

We read the correct sentences using Google Text-to-Speech (TTS) to native Korean speakers
and let them dictate the sentences they hear on crowd-sourcing platforms. The demo page for the platform that we used is shown above.

We designed our method to deliver the correct sentences to the audience in oral
because a written form may interfere the writing behavior of the audience.
As a result, we collected 51,672 transcribed sentences.

\noindent \textbf{Filtering.}
Not all transcribed sentences contain grammatical errors.
We filtered out transcribed sentences which do not contain a grammatical error by the following criteria:
\begin{enumerate}
    \item If a correct sentence and its transcription are exactly same,
    \item or if the differences between two sentences fall into any of the followings:
    \begin{itemize}
        \item a punctuation,
        \item related to a number,
        \item a named entity.
    \end{itemize}
\end{enumerate}

A punctuation is not read by TTS.
A number has multiple representations, e.g., ``1'' in Arabic numeral and ``일'' in Korean alphabet.
Finally, we excluded transcribed sentences that are too short compared to the original correct sentence.

\noindent \textbf{Why do native speakers mostly make spacing errors, while there are almost no errors on the learner corpus?}
Unlike English which generally requires a space between words, Korean often combine words without a space, depending on the context. The word spacing rule is very irregular with lots of exceptions. Consider the example 이옷은 (thiscloth) -> 이 옷은 (this cloth). In Korean, a sentence with incorrect word spacing is still comprehensible to some degree, thus people often don't strictly follow the word spacing rules. Incorrectly spaced sentences are accepted as long as they do not crucially affect readability, making word spacing rules even more difficult to memorize. On the other hand, Korean language learners may be more aware of accurate spacing due to their focus on language learning, and it is also likely that learners make other types of grammatical errors as frequent as spacing errors, which makes word spacing error much less dominant.

\subsection{\klang}
\label{appendix:lang8}
We first extract incorrect-correct Korean dataset pair from the raw Lang-8 corpus. Then, we extract all pairs that contain Korean letters and preprocess the corpus to obtain (orignial, corrected) pairs. We apply various post-processing techniques into original raw Lang-8 corpora. Those techniques include:
We discard pairs which:
\begin{itemize}
    \item token length (when tokenized by kobart tokenizer) is longer than 200, since it consisted of meaningless repetition of words or numbers.
    \item contains language other than English, Korean, or punctuations, such as arabic or japanese characters.
    \item length of one token (splitted by space) is bigger than 20, since the sentences doesn't make sense by manual inspection.
    \item contains noisy words such as 'good', 'or', '/' inside.
    \item Doesn't consist of original <-> corrected pairs.
    \item length of each sentence is at least longer than 2. (naive length, not tokenized length)

\end{itemize}

We also compute the ratio ($r_t$) of the number of tokens of the post-edit to the pre-edit ($n_{t,pre}$, $n_{t,post}$).
Similarly, we compute the ratio ($r_l$) of the lengths.
Then, we retain the (pre-edit, post-edit) pairs satisfying the following conditions and discard the others:
1) $0.25<r_t<4$,
2) $0.5<r_l<1.25$,
3) $\min(n_{t,pre}, n_{t,post}) > 5$, and
4) the length of the longest common subsequence is greater than 10 characters.

Then, we modify each sequence by deleting the traces of unneeded, additional proof marks.  Therefore we discard phrases which is inside brackets. Those indicate the subsequence SEQUENCES inside the text such as (SEQUENCES), \{SEQUENCES\}, <SEQUENCES>, or [SEQUENCES]. We discard them along WITH the brackets.
In a similar context, there was multiple repetition of partiuclar tokens, such as "안녕 홍대    !!!!!!~~~????".  so we shortened repeated patterns and make it appear only once. Those special tokens include [' ', '!', ';', '?', '\~', '>', '\^', '+', 'ㅠ', 'ㅜ', 'ㅋ']. After applying this, the original sentence is converted into "안녕 홍대 !~?"

After this step, we filter sentences by leaving out only those whose jamo\_levenshetein distance which is dicussed in Appendix \ref{appendix:jamo_lev}.\footnote{It is a morpheme-level levenshtein distance for Korean.} is smaller than 10. Pairs whose levenshtein distance is bigger than this threshold is likely to contain pairs that are not grammatical edits, but rather rephrases or additional explanations.
Lastly, we retain pairs whose original and corrected pairs are unique and original and corrected sentences are not the same (there must be at least one edit).

After this step, there are 109,560 sentence pairs in this corpus.
Full details about the modifying and filtering functions for lang8 are going to be opensourced, for reproducibility for everyone.

\subsubsection{Jamo\_levenshtein Distance}
\label{appendix:jamo_lev}
Levenshtein distance \cite{levenshtein1966binary} are computed between the pre-edit and the post-edit sentences.
We compute the distance in morpheme-level and normalize it by the sentence lengths as follows:
\begin{equation}
    \small
    ||LD|| = \frac{LD(s_{pre} , s_{post})}
         {\min(|s_{pre}|, |s_{post}|) \log_{20} \min(|s_{pre}|, |s_{post}|) }
\end{equation}
\noindent where $s_{pre}$ and $s_{post}$ denote pre-edit and post-edit, $|s|$ is length of sentence $s$,
and $LD(\cdot, \cdot)$ is Levenshetein distance, and $min(\cdot, \cdot)$ is minimum value between two arguments.
In other words, the jamo\_Levenshetein Distance between pre-edit and post-edit is normalized by their sentence length of the shorter sentence, resulting in a smaller normalization effect for longer sentence \cite{grundkiewicz2014wiked}. We use an existing implementation\footnote{\url{https://github.com/lovit/soynlp/blob/503eaee28799e9a3baf01483c6fc59e0db524fa3/soynlp/hangle/_distance.py}} which is a function inside python library called soynlp.

\subsection{\klearner}
\label{appendix:klearner}
The original corpus is a set of XML files with multiple tutors' tags and corrections to the errors of Korean learner essays. 

\subsubsection{Original XML format of the NIKL learner corpus}

\label{appendix:klearner_format}

The NIKL learner corpus consists of correction edits classified with individual tags: some of them are: (1)The position of the error(morph from-to), (2)Morpheme-level suggestions(edits) to the error(Proofread), (3)The granularity level:whether it is replacement, insertion, deletion, and so on(ErrorPattern), (4)The level of the error(ErrorLevel), (5)The role of the error in a sentence(ErrorArea), or (6)Whether it is a written or spoken language. Below are examples of the original XML dataset:
\begin{tiny}
\begin{verbatim}
Example Sentence of korean learner corpora (filetype: xml)
<s>오후 5시 반에 집에 들었어요.</s>
    ...
  <LearnerErrorAnnotations>
     <word>
      <w>들었어요.</w>
      <morph from="157" subsequence="1" to="162" wordStart="Start">
         <Proofread pos="VV">들어오</Proofread>
         <ErrorArea type="CVV" />
         <ErrorPattern type="REP" />
      </morph>
      <morph from="157" subsequence="2" to="162" wordStart="None">
         <Proofread pos="EP">았</Proofread>
      </morph>
      <morph from="157" subsequence="3" to="162" wordStart="None">
         <Preserved>어요</Preserved>
      </morph>
      <morph from="157" subsequence="4" to="162" wordStart="None">
         <Preserved>.</Preserved>
      </morph>
     </word>
  </LearnerErrorAnnotations>
\end{verbatim}
\end{tiny}
The details of how we interpret and merge edits are explained at Appendix \ref{appendix:klearner_merging}.

\subsubsection{Manual refinement step}
\label{appendix:klearner_manual}
As explained at Section \ref{sec:klearner}, some XML files had empty edits, missing tags, and inconsistent edit correction tags depending on annotators. Of all the possible tags (Appendix \ref{appendix:klearner_format}), it was common that not all ErrorArea, ErrorLevel, and ErrorPattern tags were present for each edit. Therefore we conduct a refinement step to ensure the quality of the dataset.
We process the NIKL learner corpus by the following steps: First, Merge all XML files into a single corpus. Then, we discard sentences with no or inconsistent proofread tags by \textit{manual} inspection. For example, there were datasets labeled as "DELETE" for the proofread tags, where the place was originally meant to be the place for morpheme-level edits. We discard those datasets. Since the grammatical aspects of handling written and spoken languages are different, we discard datasets tagged as spoken language and leave only written language. Lastly, we validate the consistency of the types of errors tagged by different tutors and leave out only valid annotations.
After this step, we build a corrected sentence from the original sentence and morpheme-level corrections by merging morpheme-level syllables into Korean words (Appendix \ref{appendix:klearner_merging}).

\subsubsection{The merging process from Korean orthography guidelines}
\label{appendix:klearner_merging}
In order to build corrected word-level sentences by the NIKL learner corpus, we need to apply Korean orthography guidelines since the annotations are originally decomposed in morpheme-level. We explain in detail about the rules below:
\begin{itemize}
    \item Section 18-6: When end of stem "ㅂ" is transformed to "ㅜ", write as transformed even it's against the guideline.\footnote{\url{https://kornorms.korean.go.kr/regltn/regltnView.do?regltn_code=0001&regltn_no=178\#a238}}
        % "어간의 끝 ‘ㅂ’이 ‘ㅜ’로 바뀔 경우, 그 어간이나 어미가 원칙에 벗어나면 벗어나는 대로 적는다." 
    \item Section 34: When stem ends with "ㅏ, ㅓ", using ‘-아/-어, -았-/-었-’ is harmonizing then write as it abbreviated.\footnote{\url{https://kornorms.korean.go.kr/regltn/regltnView.do?regltn_code=0001&regltn_no=178\#a254}}
        % "모음 ‘ㅏ, ㅓ’로 끝난 어간에 ‘-아/-어, -았-/-었-’이 어울릴 적에는 준 대로 적는다." 
    \item Section 35: When stem ends with "ㅗ,ㅜ", ‘-아/-어, -았-/-었-’ is harmonizing and abbreviated to "ㅘ/ㅝ,왔,웠", then write as it abbreviated.\footnote{\url{https://kornorms.korean.go.kr/regltn/regltnView.do?regltn_code=0001&regltn_no=178\#a255}}
        % "모음 ‘ㅗ, ㅜ’로 끝난 어간에 ‘-아/-어, -았-/-었-’이 어울려 ‘ㅘ/ㅝ, '으로 될 적에는 준 대로 적는다." 
    \item Section 36: When "-어" is next to "ㅣ" and abbreviated to "ㅕ" then write as it abbreviated.\footnote{\url{https://kornorms.korean.go.kr/regltn/regltnView.do?regltn_code=0001&regltn_no=178\#a256}}
            % "‘ㅣ’ 뒤에 ‘-어’가 와서 ‘ㅕ’로 줄 적에는 준 대로 적는다." 
\end{itemize}
We implemented the above Korean orthography guidelines and applied it to every sentence tokens gathered.
These method provided sufficient coverage to handle all morpheme-level corrections of corpora. We will open-source the code. But there were exceptions and uncovered cases, so in case you want to build another corpora or utilize the code to merge morphemes into words, you may want to implement more Korean orthography guidelines on our code.
We now take an example and show how we actually merged the morphemes. For example, the above XML file (Appendix \ref{appendix:klearner_format}) contains correction annotations about a morphic change from "들었어요."(meaning : came back, mis-spelled) to "들어오"+"았"+"어요".
In this case, stem "ㅗ" in "들어오" must be harmonized with ending "았" (by Korean orthography guideline, section 35).
So "들어오"+"았" must be abbreviated to "들어왔".
To handle these abbreviations, We followed these step:
\begin{itemize}
    \item join all annotations.(i.e. "들어오"+"았"+"어요"+"." = "들어오았어요."
    \item decompose all tokens to syllables (i.e. "들" is decomposed to [ㄷ,ㅡ,ㄹ] and so on)
    \item if syllable sequence applicable to abbreviation rules, then merge.(i.e. decomposed syllable sequence (ㅇ,ㅗ,None),(ㅇ,ㅏ,ㅆ) comform with Section 35.)
    \item repeat until nothing to apply
\end{itemize}

\section{KAGAS Development Details}

\subsection{Brief introduction to Korean language}
\label{appendix:intro_kor}
The current orthographical practice of Korean writing system, Hangul (한글), was established by the Korean Ministry of Education in 1988. One prominent feature of the practice is morphophonemic. This indicates that a symbol is the binding of letters consisting of morpheme-based syllables. For instance, though 자연어 in `natural language' is pronounced as 자여너 \textipa{[t\textctc a.j2.n2]}, it should be written as 자연어 since each of 자연 `natural', and 어 `language' is a morpheme with one or two syllables. Words, or Eojeol (어절) are formed by both content and functional morphemes in general. They are basic segments for word spacing in Korean. The rules for the word spacing are also described in the orthography guidelines, however, they are often regarded as complex ones for native Korean speakers \cite{OUNHCHR2014}.

In the view of linguistic typology, as mentioned, Korean is an agglutinative language in that each morpheme encodes a single feature. This turns out that the language has rich morphology such as various particles and complex conjugation forms. The example in (1) shows that each particles attached to a noun indicates a case marker such as nominative, accusative and the others. Furthermore, the affixes attached to a verb stem serve as functional morphemes pertaining to tense, aspect and mood. Another distinction of the language is that pro-drop or zero anaphora is abundant, which is common in morphologically rich languages \cite{tsarfaty2010statistical}. Particle omission is also frequent in colloquial speech \cite{lee2012annotating}. These linguistic characterisitcs are different from the ones of fusional languages such as English and German where a concatenated morpheme has multiple features in usual \cite{comrie1989language, vania2017characters}.

\begin{exe}
  \ex
  \gll  수지-가 한나-에게 우체국-에서 편지-를 보내-는 중-이-라고 말-했-습-니-다 \\
	      Suzy-\textsc{nom}  Hannah-\textsc{dat} post.office-\textsc{loc} letter-\textsc{acc} send-\textsc{prs} currently.doing-\textsc{adj}-\textsc{quot} say-\textsc{pst}-\textsc{pol}-\textsc{ind}-\textsc{decl} \\
  \glt  `Suzy said that she was sending a letter to Hannah at the post office.'
\end{exe}

The word order of Korean is relatively free. While the canonical word order of the language is Subject-Object-Verb (SOV), it is possible to change the positions of the words in a sentence. However, corpus and psycholinguistic studies have reported that the preferred order exists for adverbs with the conditions related to the meaning of the verb \cite{shin2007corpus} and the specific types of the adverbs such as the time and place ones \cite{Nam2018}. Nevertheless, Korean speakers allow various word orders when comprehending and producing sentences in general.

\subsection{Benefits of using KAGAS}
\label{appendix:benefits_of_kagas}
\begin{table}
\centering
\resizebox{0.49\textwidth}{!}{
\begin{tabular}{@{}lll@{}}
\toprule
 & Human annotations & KAGAS \\ \midrule
Schema &
  \begin{tabular}[c]{@{}l@{}}Different by datasets.\\ Cannot compare\end{tabular} &
  \begin{tabular}[c]{@{}l@{}}Provides a \\ unified schema\end{tabular} \\ \midrule
Decisions &
  \begin{tabular}[c]{@{}l@{}}Random\\ Differs by annotators\end{tabular} &
  \begin{tabular}[c]{@{}l@{}}Deterministic\\ More trustworthy\end{tabular} \\ \midrule
Time/price cost &
  \begin{tabular}[c]{@{}l@{}}Experts needed.\\ Expensive\end{tabular} &
  \begin{tabular}[c]{@{}l@{}}No cost, can\\ instantly get output\end{tabular} \\ \bottomrule
\end{tabular}}
\caption{Benefits of KAGAS over human annotations.}
\label{tabs:kagas_vs_human}
\end{table}
Table \ref{tabs:kagas_vs_human} shows the benefits of using an automatic system over human annotation. Making a fully human annotated dataset resource for Grammatical Error Correction is quite difficult and costly, and an automated version of it (KAGAS) could be a great alternative, and could even overcome many disadvantages of human annotation. Therefore, we emphasize here again the advantages of automatic error type correction system.
\begin{itemize}
    \item KAGAS provides a unified schema for all Korean parallel datasets. In contrast, error types by human annotations are different by datasets, and thus hard to compare.
    \item KAGAS uses a deterministic, trustworthy decision on assigning error types, where it could be random or different by annotators for human annotations.
    \item KAGAS can be applied with no cost, and instantly get output while it takes a lot of time, money, and effort to hire experts for annotation and validate them. This particularly becomes a great advantage for datasets used for training neural models where the dataset size is often too large to conduct high-quality human annotation, and on other languages than English where experts are very expensive and difficult to hire.
\end{itemize}

\subsection{More examples of Korean Error Types}
\subsubsection{INS \& DEL}
\label{appendix:ins_del}
In Korean, one can omit the subject or the object in a sentence depending on the previous context, since it is a discourse-oriented language. In these cases, sentence with DEL and INS edits and also without DEL and INS edits are both grammatically correct. There were also cases of INS or DEL which are edits of unnecessary modifiers. This is also a case of non-grammatical edits but rather variations to sentences. For this reason, we felt no need to divide error types further for INS\&DEL that mostly accounts for unnecessary, \textbf{non-grammatical edits}. Below are some INS \& DEL examples from lang8.txt.

Line.1825 :

음악회에 가는 것은 좋아해서 ... (Like going to musicals)

INS->``나는'' 음악회에 가는 것은 좋아해서 ... (I like going to musicals)

Line. 1909 :

날마다 ``이'' 일기에 써고 싶어요 (Want to write on this diary everyday)

DEL->날마다 일기에...(Want to write on diary everyday)

-All sentences are grammatically correct in Korean. 

\subsection{Detailed examples of Alignments and assignment of error types by KAGAS}
\label{appendix:morpheme_alignment_examples}
We add some examples that describe how KAGAS assigns word-level POS error types by morpheme-level edits. Note that POS tagging is conducted by the kkma POS tagger.

\begin{itemize}
    \item Word-level \textbf{ins}ertion
        \begin{itemize}
        \item 음악회에 가는... => \textbf{나는} 음악회에 가는 ..
        \item “나는” is inserted
        \item INSERTION
        \end{itemize}
    \item Morpheme-level \textbf{del}etion
        \begin{itemize}
        \item 소풍(NNG) + 을(JKO) =>소풍(NNG)
        \item 을(JKO) is deleted, thus labeled as PART(JKO is grouped to PART)
        \end{itemize}
    \item Morpheme-level \textbf{ins}ertion
        \begin{itemize}
        \item 유학(NNG) => 유학(NNG) + 러(NNP)
        \item 러(NNP) inserted, thus labeled as Noun(NNP -> PART)
         \end{itemize}
    \item Morpheme-level \textbf{sub}stitution
        \begin{itemize}
        \item ``싶습니다'' => ``싶어합니다''
        \item tokenized by POS tagger as ``싶''+``습니다'' => ``싶''+``어''+``하''+``ㅂ 니다''
        \item 싶(VXA) + 습니다(EFN) -> 싶(VXA) + 어(ECD) + 하(VV) + ㅂ 니다(EFN)
        \item (EFN->ENDING) to (ECD->ENDING, VV->VERB, EFN->ENDING)
        \item (ENDING) -> (ENDING, VERB)
        \item sum((ENDING), (ENDING,VERB)) -> (VERB, ENDING)
        \item If we aggregate all POS included in this edit, we get (VERB, ENDING), and therefore is labeled into ``CONJ''.
        \end{itemize}
\end{itemize}
Please refer to software:KAGAS/pos\_granularity.py for full mapping of kkma POS tags grouped into our error types.

\subsection{On assignment of single representative Error Types instead of multiple Error Types per a single edit}
\label{appendix:why_not_multiple_errortypes}
Defining error types that don't overlap with one another in the first place would be optimal, but unfortunately, defining meaningful error types for Korean that are mutually exclusive is almost infeasible. (Appendix \ref{appendix:klearner_format}: The NIKL corpus tagged
morpheme-level error types in 3 levels: the position of error types(오류 위치), ErrorPattern(오류 양상), and ErrorLevel(오류 층위).) Similarly, we want to clarify that the
current implementation of KAGAS(software:KAGAS/scripts/align\_text\_korean.py\#L404) has the ability to output all candidates of error type classifications(in formal aspect (INS/DEL/SUB), the POS of the edit, and the nature and scope of edit(SPELL/SHORT/The rest)). Currently, it is aggregated to a single error type, in the order of pre-defined priorities. While KAGAS can be easily extended to output multiple error types for a single edit, human evaluation and error type distribution analysis becomes much more complicated if we evaluate all possible error types per edit. For simplicity and clarity (and to make a deterministic reliable system), we decided to assign priorities and conduct human evaluation only on the highest priority error types. Please note that other works that extend ERRANT onto other languages also assign single error types to each edit \cite{naplava-etal-2022-czech}, \cite{sonawane-etal-2020-generating}, \cite{katinskaia-etal-2022-semi}.

\subsection{The granularity of Error Types}
\label{appendix:granularity}
Our primary goal on building KAGAS was to correctly classify error types in as much coverage as possible, while the human evaluation of KAGAS output is reliable enough. The first version of KAGAS was made after referring to the Korean orthography guidelines and other related work, and adjusting them into the ERRANT error types. It first contained a more diverse set of error types, with multiple error types assigned per an edit (e.g. SUB:VERB:FINAL\_ENDING, SUB:VERB:DERIVATIONAL\_ENDING, SUB:PARTICLE:OBJECTIVE, or INS:PUNCT). However, we noticed that there were 2 issues that prevented the practical and reliable use of the first version of KAGAS, and fixing these problems led to the current version described in the paper. First, the accuracy of kkma (POS tagger) was not good enough to ensure good quality of error types described previously in much detail, which is something that is beyond the scope of this work (We believe that the improvement of a POS tagger will enable KAGAS to define a more detailed error type classification with high reliability). Second, we could not perform reliable human evaluation with fine-grained error types. For reliable human evaluation we needed at least 26 samples per an error type - 13 in Kor-Native and 13 in Kor-Learner + Kor-Lang8 - to conduct a reliable human evaluation. Therefore, to ensure the quality of classification by KAGAS, error types without sufficient samples were aggregated into higher categories of similar groups or left as unclassified (at software:KAGAS/edit- extraction/pos\_granularity.py).

\section{Implementation Details}
\subsection{Kkma POS Tagger}
\label{appendix:kkma_pos_tagger}
 We use the Konlpy wrapper for Kkma Korean POS tagger,\footnote{\url{https://konlpy-ko.readthedocs.io/ko/v0.4.3/}} to tag Part-Of-Speech information in a given sentence. We chose to use Kkma because it had the most diverse POS tags \footnote{\url{http://kkma.snu.ac.kr/documents/index.jsp?doc=postag}} among the konlpy POS taggers. However, Kkma fail to recover to the original form of a sentence after the output of POS tagging. Kkma outputs morpheme-level tags, and it erases whitespaces from the original input sentence. Therefore, recovering whitespaces after processing a sentence by Kkma is necessary, along with aggregating morpheme-level tags into word-level. We solve this issue by utilizing morpheme-level alignment for Korean.
\subsection{Defining the priorities between error types }
\label{appendix:kagas_error_type_priorities}
We wanted our system to be highly reliable and clear given the current available resources. Therefore, we prioritized classifying frequent, orthographic error types over POS classification.

After the output of the edit extraction, We use the allSplit method and merge mulitple edits as one edit of word space and word order errors. For detection spell errors, we explained earlier that we use the Korean spellchecker dictionary.\footnote{https://github.com/spellcheck-ko/hunspell-dict-ko} Note that words that are proper noun is likely to be not included in the Korean dictionary, so spell errors are defined in a more narrower sense than it is currently thought of. We defined edits as spell errors only when original span wasn't inside the korean dictionary, but after editing, the edited word is inside the Korean dictionary. Therefore, corrections on proper nouns are treated as correct when there are classified as NOUN errors, not SPELL errors.  

There were sometimes edits that could be both classified by one or more error types. For example, and insertion edit that added punctuation can be both classified as "INS" edits or "PUNCT" edits. In order to avoid this ambiguouity, we set the priority between edits. The priority is as follows.
\begin{itemize}
\item INS \& DEL > the rest
\item WS > WO > SPELL > SHORT > PUNCT > the rest
\end{itemize}

We informed this to participants for human evaluation to evaluate ambiguous edits on this priority.
% Explanations that korean learners are morepheme-level, explanations for why we should make word-level edits and what efforts and merging rules we used to make it word-level edits should be included.
For Korean-specific linguistically aligned alignment, we computed similar with the English alignment system, but we defined Korean lemma cost using the soylemma's lemmatizer, and we defined the Korean content pos as NNG, NNP, NNB, NNM, NR, NP, VV, VA, VXV, VXA, VCP, VCN, MDT, MDN, MAG, MAC out of full pos tags for korean.\footnote{http://kkma.snu.ac.kr/documents/index.jsp?doc=postag}

\subsection{Qualitative analysis on user evaluation.}
\begin{figure}[!ht]
\centering
{
\includegraphics[width=0.9\linewidth]{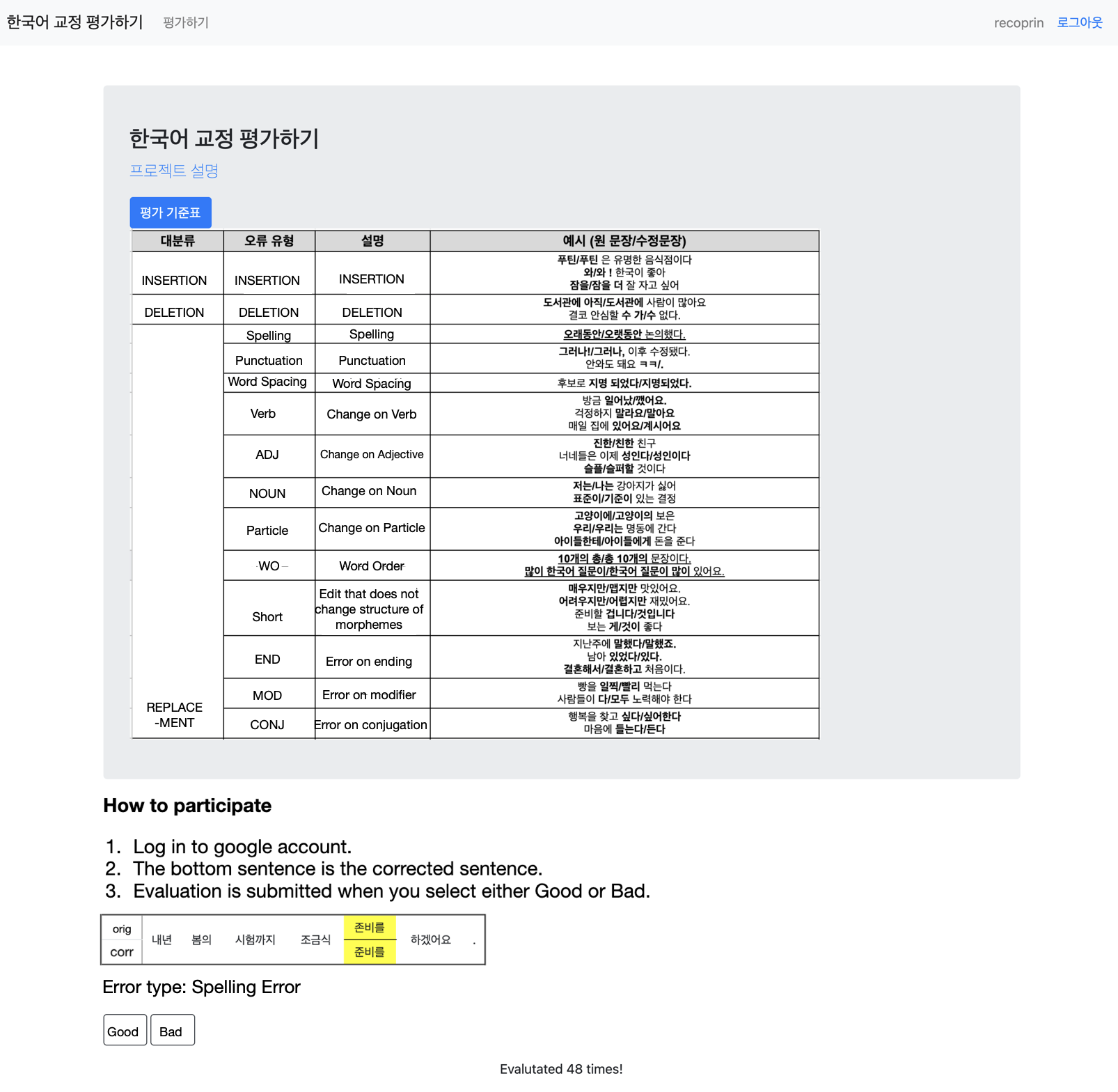}
}
\caption{Demo that we used for KAGAS system evaluation. Translated into english.}
\label{figs:human_eval}
\end{figure}
\subsubsection{Evaluation Interface}
Figure \ref{figs:human_eval} shows the evaluation demo interface that we used for human evaluation. We gave the full list of error types and made the evaluators to mark either 'good' or 'bad' about the error type classification.
\label{appendix:analysis_by_user_evaluation}
\subsubsection{About low-performing cases}
Overall, the participants evaluated error types that could easily be identified by their forms with a higher proportion of 'good', and error types that relates to the POS tags as 'bad'. After manual inspection of edits that were classified as 'bad' by the Korean experts, we found that most of them were due to the limitations of the POS tagger. Most of the times the POS tagger fail to tag the correct POS for edit words, especially when there is a spelling error inside a word, or it is a pronoun. This explains why acceptance rates for POS-related error types had lower scores, for example, ADJ, NOUN, VERB, or CONJ. Also, after the main evaluation, we additionally asked the participants to classify edits that were marked as UNK, edits which KAGAS was unable to classify it to any error types. The participants classified most of the UNK edits as spelling errors. Since there are a lot of inflectional forms for a word for Korean, current dictionary-based spellchecker fail to identify all SPELL error edits. Therefore, we believe that KAGAS will benefit from the improvement of the Korean POS tagger and spell checker.

\subsubsection{About selection of sentences for evaluation}
For simplicity and clarity for annotators, we selected sentences with a \textbf{single} edit for each error type from our dataset for human evaluation. One concern could be that there could be a selection bias - straightforward cases could be selected for evaluation. We would like to first clarify that our 14 error types are entirely defined by local edits. In other words, the error type classification output of KAGAS is not affected by adjacent words or sentence structure (POS tagging is performed word-wise, and INS/DEL edits are not divided further). Therefore, we carefully argue that the validity of KAGAS is not affected by the number of edits and thus sentences with one edit can sufficiently represent the entire data, since the goal of human evaluation is to evaluate whether KAGAS correctly classifies word-level edits.

\section{Experimental Details}
\label{appendix:exp-details} 

\subsection{Details of Experimental Settings}
We used a computational infrastructure which have 4 core CPU with 13GB memory and with one GPU(NVIDIA Tesla V100). All reported models are run on one GPU.
We use the kobart pretrained model and kobart tokenizer. We allocate 70\% of data set to train, 15\% to test, and 15\% to valid data sets by using Python scikit-learn library, sklearn.train\_test\_split function. GLEU\cite{napoles-etal-2015-ground} scores are evaluated by the official github repository \footnote{https://github.com/cnap/gec-ranking}, and $M^2$ scores \cite{dahlmeier-ng-2012-better} are also evaluated on the official repository.\footnote{https://github.com/nusnlp/m2scorer}

\begin{table*}[t]
\tiny
\centering
\setlength\tabcolsep{4pt}
\begin{threeparttable}
\resizebox{\textwidth}{!}{
\begin{tabular}{|c|c|c|c|c|c|c|c|c|c|c|c|c|c|c|c|c|c|c|c|c|}
\hline
 &
  \multicolumn{5}{c|}{\klearner{}} &
  \multicolumn{5}{c|}{\knative{}} &
  \multicolumn{5}{c|}{\klang{}} &
  \multicolumn{5}{c|}{\kunion{}} \\ \hline
 &
  \multirow{2}{*}{GLEU} &
  \multicolumn{3}{c|}{$M^2$} &
  \multirow{2}{*}{\begin{tabular}[c]{@{}c@{}}total\\ time\end{tabular}} &
  \multirow{2}{*}{GLEU} &
  \multicolumn{3}{c|}{$M^2$} &
  \multirow{2}{*}{\begin{tabular}[c]{@{}c@{}}total\\ time\end{tabular}} &
  \multirow{2}{*}{GLEU} &
  \multicolumn{3}{c|}{$M^2$} &
  \multirow{2}{*}{\begin{tabular}[c]{@{}c@{}}total\\ time\end{tabular}} &
  \multirow{2}{*}{GLEU} &
  \multicolumn{3}{c|}{$M^2$} &
  \multirow{2}{*}{\begin{tabular}[c]{@{}c@{}}total\\ time\end{tabular}} \\ \cline{1-1} \cline{3-5} \cline{8-10} \cline{13-15} \cline{18-20}
 &
   &
  Pre. &
  Rec. &
  $F_{0.5}$ &
   &
   &
  Pre. &
  Rec. &
  $F_{0.5}$ &
   &
   &
  Pre. &
  Rec. &
  $F_{0.5}$ &
   &
   &
  Pre. &
  Rec. &
  $F_{0.5}$ &
   \\ \hline
\begin{tabular}[c]{@{}c@{}}Self-Scores\\[5pt]  \end{tabular} &
  25.90 &
  1 &
  0 &
  0 &
  - &
  25.92 &
  1 &
  0 &
  0 &
  - &
  19.38 &
  1 &
  0 &
  0 &
  - &
  21.66 &
  1 &
  0 &
  0 &
  - \\ \hline
\begin{tabular}[c]{@{}c@{}}Hanspell \\[5pt] \end{tabular} &
  29.75 &
  24.17 &
  5.37 &
  14.21 &
  7 &
  57.97 &
  81.16 &
  47.80 &
  71.22 &
  4 &
  22.35 &
  29.09 &
  8.51 &
  19.61 &
  71 &
  28.46 &
  36.30 &
  11.49 &
  25.35 &
  69 \\ \hline
\begin{tabular}[c]{@{}c@{}}KoBART \\[5pt] \end{tabular} &
  46.94 &
  43.95 &
  26.35 &
  38.76 &
  1hr 4min &
  69.37 &
  75.07 &
  56.81 &
  70.53 &
  38min &
  28.57 &
  37.69 &
  12.64 &
  26.96 &
  3hr 39min &
  34.07 &
  44.33 &
  15.33 &
  32.13 &
  5hr 6min \\ \hline
\begin{tabular}[c]{@{}c@{}}KoBART +\\ \kunion{}\end{tabular} &
  44.66 &
  51.94 &
  23.55 &
  41.83 &
  1hr &
  61.64 &
  83.95 &
  48.55 &
  73.25 &
  39min &
  28.51 &
  38.53 &
  12.72 &
  27.40 &
  3hr 38min &
  - &
  - &
  - &
  - &
 - \\ \hline

\end{tabular}}
\end{threeparttable}
\label{tabs:overall_experiment_test}
\caption{We also report the evaluation results on valid sets which has the highest GLEU score. KoBART outputs are averaged from outputs of 3 different seeds. Here, we also report the total training time for 10 epochs, in the total time section.}
\end{table*}

\subsection{Tokenizer issue on punctuation space recovery}
\label{appendix:tokenizer_issue}
Below is an example of the encoded and decoded outputs of the tokenizer.
\begin{verbatim}
>>> orig_text = "이게 뭔가요 왜 안돼요 ? ."
>>> orig_tokens = tokenizer.encode(orig_text)
>>> orig_tokens
[17032, 20156, 11900, 14851, 14105, ... 17546]
>>> decoded_text = tok.decode(orig_tokens)
>>> decoded_text
'이게 뭔가요 왜 안돼요?.'
>>> orig_text == decoded_text
False
\end{verbatim}
We can see that spaces between punctuations and word disappeared from the decoded text, thus making it different from the original raw text. For this reason, we conduct the KAGAS experiments and report error type distributions on each datasets by raw text, but use the decoded version on evaluating model outputs. Full error type distributions on both raw text and decoded text is described on Table \ref{appendix:full_errortype_counts}.

\noindent \textbf{BART.} We use BART-base architecture, having the number of trainable parameters are 123M. we evaluate on the model which scores highest in GLUE scores. We first started with hyperparameters that were used to fine-tune BART on CNN-Dailymail task.\footnote[1]{\url{https://github.com/pytorch/fairseq/blob/master/examples/bart/README.summarization.md}} Among those parameters, we experimented on different dropout rates, and on the learning rates. 
We use the BartForConditionalGeneration structure by the huggingface library.\footnote{https://huggingface.co/transformers/} When generating, number of beams that we used for beam search is fixed to 4.

\subsection{About pre-training with wikipedia dataset.}

Since GEC suffers from lack of plentiful parallel data, we also tried to pre-train our model on Wikipedia edit pairs \cite{lichtarge-etal-2019-corpora} with a learning rate of 1e-05, and then fine-tune for 10 epochs on each individual datasets. 
However, we found out that KoBART is already a very strong pre-trained model, and the benefit from Wikipedia edit pair training is small. Therefore, we decided not to use the Korean Wikipedia edit pairs on our baseline experiment.
\subsection{Further analysis on model outputs.}
\label{appendix:model_output_analysis}

\begin{figure*}[!h]
\centering
\subfloat[Hanspell]{\includegraphics[width=0.48\textwidth]{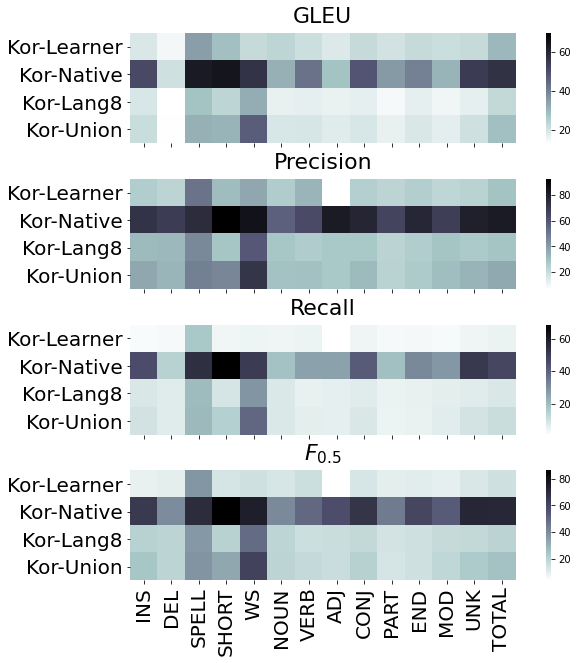}\label{fig:main_hanspell_heatmap}}
\quad
\subfloat[KoBART]{\includegraphics[width=0.48\textwidth]{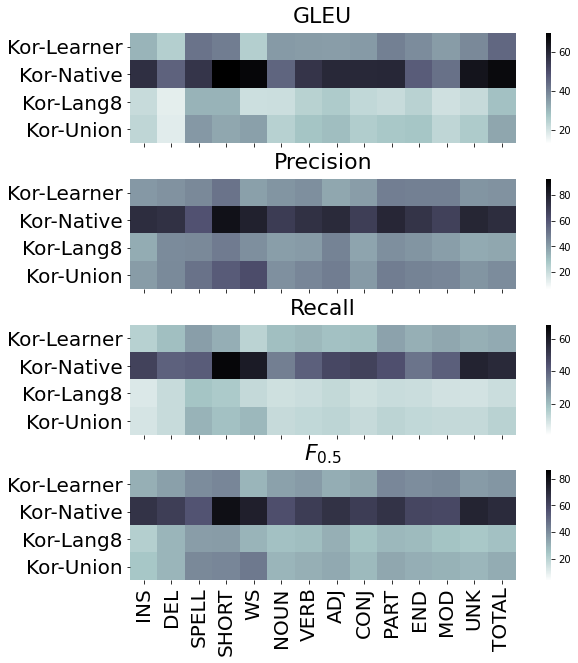}\label{fig:main_kobart_heatmap}}
\caption{Heatmap illustration of generation output on test set of (a) Hanspell  and (b) KoBART error types. We leave out WO and PUNCT due to the lack of examples in the test set. We can see that the scores of KoBART are similar over all error types, while Hanspell scores are biased toward word spacing (WS) and spelling (SPELL). Full values are in Appendix Table \ref{tabs:errortype_scores_on_everything}.}
\label{fig:main_heatmap}
\end{figure*}

\begin{table}[!t]
\centering
\small
\setlength\tabcolsep{2pt}
\begin{tabular}{@{}cccc@{}}
\toprule
 & \textbf{\klearner} & \textbf{\knative} & \textbf{\klang} \\ 
 \midrule
\# r & .50 & .46 & .03 \\ 
\midrule
p & .06 & .09 & .92 \\ 
\midrule
Significant & O & O & X \\ 
\bottomrule
\end{tabular}
\caption{Correlation between error type proportions(\%) with respect to GLEU scores of KoBART + individual. The correlation is significant (p < .1) for \klearner{} and \knative.}
\label{tab:correlation_proportion}
\end{table}
\paragraph{Accuracies improve linearly with the proportion of the training dataset.} 
\knative{} scores notably high on WS, while \klearner{} scores poorly. According to Figure \ref{fig:errortype_distribution}, \knative{} has a large proportion of WS and \klearner{} have only a small fraction. The same trend applies for PART on \klearner{} datasets, compared with \klang{}. Table \ref{tab:correlation_proportion} is obtained by the indiviual error type proportion with respect to the GLEU scores shown in Figure \ref{fig:main_heatmap}. According to Table \ref{tab:correlation_proportion}, there is a positive correlation between the distribution of error types and the individual performance for \klearner{} and \knative{}.
This means that when making a GEC model, training dataset distributions should differ in relation to what type of error types one wants to have high performance on. For example, if a model that performs well on ADJ errors is needed, \klearner{} dataset should be utilized, and if a model that corrects WS errors very well is needed, the \knative{} dataset should be used, and if one need to correct informal errors from KFL learners, using \klang{} would be the best, while using \knative{} would be better to correct native speaker errors. Therefore, we believe all three datasets have their own purpose, and we provide them as separate three datasets without unifying them.
\begin{figure}[!h]
\centering
{
\includegraphics[width=0.49\linewidth]{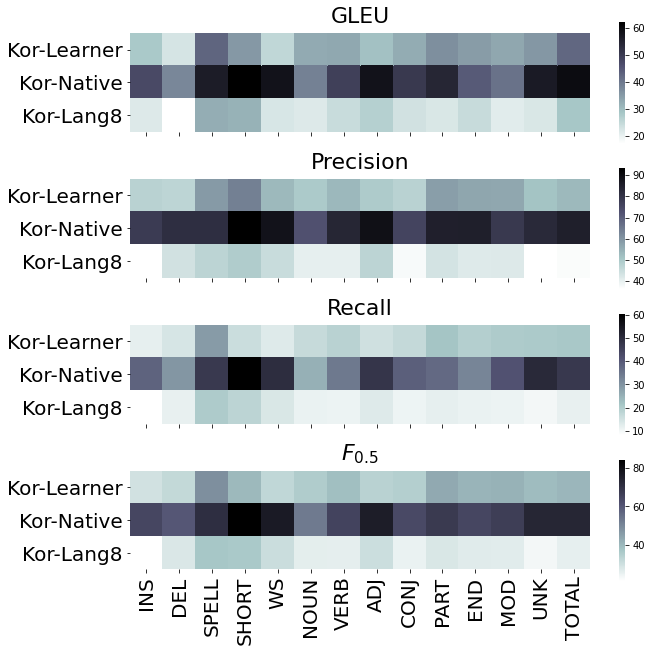}
}
\caption{Heatmap illustration of error types generated by KoBART + Union and individual dataset fine-tuning. Note that the scoring range is different from that of Figure \ref{fig:main_heatmap_valid}.}
\label{fig:union_heatmap}
\end{figure}

\paragraph{Comparison with KoBART and KoBART + \kunion{}}
The results on KoBART + \kunion{} is the results from the model fine-tuned twice, first with \kunion{} and then with the individual dataset.
As we can see in Table \ref{tabs:experiments_overall}, there is an improvement in precision and $F_{0.5}$ scores compared with KoBART+\kunion{} (KoBART fine-tuned on \kunion{} and then fine-tuned on each individual dataset again) than KoBART (fine-tuned directly on KoBART), meaning that all three datasets can help on improving the performance of the individual datasets. Analysis for KoBART + \kunion{} on each error type distributions shows similar trends with KoBART (Table \ref{tabs:errortype_scores_on_everything}). 
\paragraph{Additional Results}
Figure \ref{fig:union_heatmap} shows the test dataset heatmap for KoBART + \kunion{}. We can see that the trends are similar with that of KoBART.
\begin{figure*}[!h]
\centering
\subfloat[Hanspell]{\includegraphics[width=0.49\textwidth]{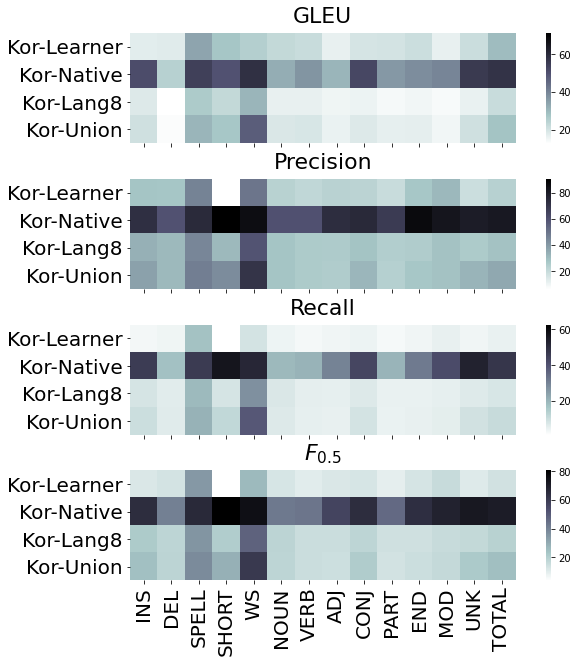}\label{fig:main_hanspell_heatmap}}
\hfill
\subfloat[KoBART]{\includegraphics[width=0.49\textwidth]{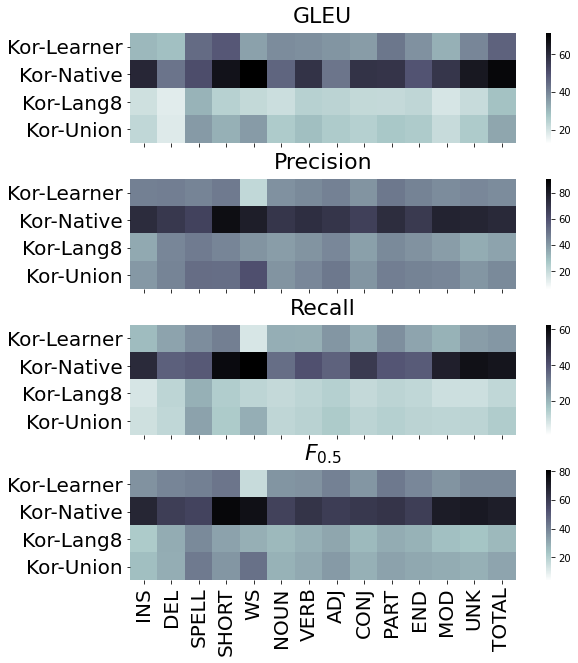}\label{fig:main_kobart_heatmap}}
\caption{Heatmap illustration of generation output of valid set by (a) Hanspell  and (b) KoBART error types. We can see that the distributions are similar to those on the test set.}
\label{fig:main_heatmap_valid}
\end{figure*}
Also, Figure \ref{fig:main_heatmap_valid} shows the valid dataset heatmap of KoBART compared with Hanspell.
The full error type scores are described in Table \ref{tabs:errortype_scores_on_everything}. 
It first shows the count of occurences of the valid datasets used for generation and making heatmap illustrations. Also, it shows the full output of scores by all datasets with all methodologies.
\paragraph{Discussion on the ability of our model to refrain from editing when no fix is necessary}
Another important aspect to grammatical error correction models would be about how these tools behave in the face of both grammatically incorrect and correct sentences. For this, we have evaluated the GEC model fine-tuned on the Kor-Native train dataset on both the source (grammatically incorrect) and target (grammatically correct) sentences of the 2,634 Kor-Native dev set. When we provided the model with grammatically correct sentences as input, the model output the exact same sentences as the original input sentence in 2184/2634 cases (82.92\%). In contrast, when we provided grammatically incorrect sentences as input, the model only preserved the 400/2634 input sentences (15.19\%) and fixed the input in 84.81\% of the cases. This shows that while the model was only trained with grammatically incorrect sentences as input, it has developed an ability to determine whether a sentence is grammatically correct or not, thus refraining from editing in such cases. We will open-source the code to run this experiment and include the results on our next revision. Since we have presented our model as a baseline, we hope that many improvements can be made for this aspect in future work, e.g., explicitly being trained to preserve correct sentences.

\begin{table*}[ht]
\centering
\begin{threeparttable}
\resizebox{0.8\textwidth}{!}{
\begin{tabular}{|c|cc|cc|cc|}
\hline
\multirow{2}{*}{Error Type(Full)} & \multicolumn{2}{c|}{\klearner{}} & \multicolumn{2}{c|}{\knative{}} & \multicolumn{2}{c|}{\klang{}} \\
             & raw   & decoded & raw   & decoded & raw    & decoded \\ \hline
INS    & 3352  & 3321      & 2004  & 1998      & 32665  & 28040     \\
DEL     & 1652  & 1629      & 806   & 642       & 30666  & 27334     \\
SPELL        & 6735  & 5642      & 3208  & 1078      & 19021  & 14506     \\
PUNCT        & 0     & 19        & 2     & 79        & 284    & 3778      \\
SHORT      & 363   & 374       & 115   & 277       & 857    & 974       \\
WS           & 108   & 158       & 15625 & 15617     & 9766   & 9653      \\
WO           & 0     & 0         & 45    & 45        & 701    & 676       \\
NOUN         & 4879  & 5039      & 1442  & 1459      & 20300  & 20473     \\
VERB         & 2456  & 2557      & 486   & 523       & 8616   & 8805      \\
ADJ    & 411   & 444       & 55    & 88        & 1657   & 1753      \\
CONJ  & 4917  & 5269      & 1699  & 2009      & 28775  & 31078     \\
PART     & 16700 & 16692     & 1164  & 1195      & 39648  & 39175     \\
END       & 7560  & 7310      & 591   & 683       & 25456  & 22495     \\
MOD     & 1035  & 1043      & 238   & 258       & 5320   & 5260      \\
UNK & 9251  & 9940      & 2495  & 3921      & 39101  & 43132     \\
TOTAL        & 59419 & 59437     & 29975 & 29872     & 262833 & 257132    \\ \hline
\end{tabular}}
\end{threeparttable}
\caption{Total count of edits on each error type on full data. Raw count is derived from KAGAS with raw dataset pairs, and Tokenized count is derived from KAGAS with tokenized->detokenized (decoded) dataset pairs using the Korean character BPE tokenizer.}
\label{appendix:full_errortype_counts}
\end{table*}

\begingroup
\onecolumn
\setlength{\tabcolsep}{3pt}
\tiny
\begin{longtable}{@{}ccccccccccccccccccc@{}}
\toprule
 &                           &           & INS   & DEL   & SPELL & PUNCT  & SHORT & WS    & WO     & NOUN  & VERB  & ADJ   & CONJ  & PART  & END   & MOD   & UNK   & TOTAL \\ \midrule
\multirow{8}{*}{\begin{tabular}[c]{@{}c@{}}Dataset \\ Count\end{tabular}} &
  \multirow{4}{*}{valid} &
  \klearner &
  450 &
  249 &
  828 &
  0 &
  65 &
  19 &
  0 &
  785 &
  394 &
  77 &
  769 &
  2464 &
  1084 &
  149 &
  1483 &
  8816 \\
 &                           & \knative  & 331   & 94    & 196   & 21     & 34    & 2336  & 4      & 218   & 74    & 14    & 312   & 198   & 92    & 50    & 633   & 4607  \\
 &                           & \klang    & 4179  & 4131  & 2092  & 604    & 160   & 1434  & 95     & 3067  & 1350  & 248   & 4863  & 5878  & 3367  & 792   & 6486  & 38746 \\
 &                           & \kunion   & 4960  & 4474  & 3116  & 625    & 259   & 3789  & 99     & 4070  & 1818  & 339   & 5944  & 8540  & 4543  & 991   & 8602  & 52169 \\
 & \multirow{4}{*}{test}     & \klearner & 517   & 205   & 879   & 3      & 54    & 30    & 0      & 797   & 408   & 63    & 744   & 2405  & 1104  & 163   & 1506  & 8878  \\
 &                           & \knative  & 304   & 108   & 156   & 12     & 38    & 2383  & 10     & 208   & 90    & 15    & 308   & 176   & 103   & 40    & 578   & 4529  \\
 &                           & \klang    & 4187  & 4164  & 2151  & 571    & 136   & 1440  & 95     & 3119  & 1323  & 225   & 4587  & 6042  & 3283  & 815   & 6521  & 38659 \\
 &                           & \kunion   & 5008  & 4477  & 3186  & 586    & 228   & 3853  & 105    & 4124  & 1821  & 303   & 5639  & 8623  & 4490  & 1018  & 8605  & 52066 \\ \midrule
\multirow{16}{*}{\begin{tabular}[c]{@{}c@{}}Hanspell\\ (valid)\end{tabular}} &
  \multirow{4}{*}{\klearner} &
  gleu &
  18.45 &
  18.50 &
  34.48 &
  0.00 &
  27.67 &
  25.77 &
  0.00 &
  23.23 &
  22.42 &
  17.02 &
  20.26 &
  20.54 &
  21.78 &
  17.08 &
  22.20 &
  29.75 \\
 &                           & prec      & 28.42 & 27.78 & 46.34 & 100.00 & 6.67  & 50.00 & 100.00 & 23.54 & 21.74 & 23.68 & 22.99 & 19.75 & 28.11 & 31.94 & 19.45 & 24.17 \\
 &                           & rec       & 3.71  & 4.36  & 17.82 & 100.00 & 1.42  & 9.30  & 100.00 & 4.55  & 3.58  & 4.09  & 4.81  & 3.18  & 4.29  & 5.60  & 4.29  & 5.37  \\
 &                           & $f_{0.5}$ & 12.18 & 13.39 & 35.10 & 100.00 & 3.83  & 26.67 & 100.00 & 12.84 & 10.79 & 12.10 & 13.09 & 9.68  & 13.31 & 16.45 & 11.40 & 14.21 \\
 & \multirow{4}{*}{\knative} & gleu      & 51.37 & 25.25 & 54.56 & 20.14  & 50.00 & 58.68 & 0.00   & 32.84 & 37.43 & 31.26 & 52.73 & 36.67 & 38.76 & 40.29 & 56.27 & 57.97 \\
 &                           & prec      & 72.78 & 60.00 & 74.44 & 55.00  & 90.48 & 85.30 & 100.00 & 60.12 & 60.34 & 73.33 & 74.94 & 68.18 & 86.76 & 82.46 & 79.77 & 81.16 \\
 &                           & rec       & 46.16 & 18.18 & 46.64 & 16.18  & 56.72 & 52.00 & 16.67  & 19.80 & 20.83 & 30.56 & 43.35 & 20.59 & 32.07 & 41.96 & 53.28 & 47.80 \\
 &                           & $f_{0.5}$ & 65.25 & 41.10 & 66.51 & 37.16  & 80.85 & 75.61 & 50.00  & 42.72 & 43.75 & 57.29 & 65.40 & 46.63 & 64.69 & 69.12 & 72.56 & 71.22 \\
 & \multirow{4}{*}{\klang}   & gleu      & 18.93 & 13.22 & 27.21 & 15.59  & 23.51 & 32.10 & 10.02  & 16.90 & 17.91 & 15.54 & 16.44 & 14.69 & 15.46 & 13.74 & 17.12 & 22.35 \\
 &                           & prec      & 33.97 & 32.04 & 46.06 & 29.82  & 34.62 & 58.39 & 22.73  & 27.45 & 25.93 & 22.52 & 29.70 & 24.80 & 26.66 & 26.03 & 26.88 & 29.09 \\
 &                           & rec       & 8.91  & 6.89  & 20.20 & 5.88   & 8.70  & 27.86 & 4.17   & 7.48  & 5.91  & 4.76  & 7.95  & 5.35  & 5.39  & 5.20  & 7.38  & 8.51  \\
 &                           & $f_{0.5}$ & 21.74 & 18.52 & 36.67 & 16.43  & 21.69 & 47.89 & 12.02  & 17.89 & 15.45 & 12.89 & 19.20 & 14.37 & 14.89 & 14.45 & 17.58 & 19.61 \\
 & \multirow{4}{*}{\kunion}  & gleu      & 21.27 & 13.85 & 31.22 & 15.75  & 27.81 & 47.80 & 9.48   & 19.37 & 19.81 & 16.72 & 18.85 & 17.28 & 17.54 & 15.39 & 21.08 & 28.46 \\
 &                           & prec      & 38.15 & 31.80 & 47.84 & 30.19  & 43.75 & 70.90 & 24.44  & 28.30 & 26.43 & 25.94 & 32.40 & 24.57 & 27.87 & 29.11 & 33.09 & 36.30 \\
 &                           & rec       & 10.66 & 7.10  & 21.16 & 6.27   & 12.38 & 39.21 & 4.47   & 7.61  & 5.98  & 5.66  & 9.12  & 5.19  & 5.60  & 6.31  & 9.81  & 11.49 \\
 &                           & $f_{0.5}$ & 25.16 & 18.76 & 38.21 & 17.12  & 29.03 & 61.04 & 12.91  & 18.34 & 15.70 & 15.12 & 21.45 & 14.07 & 15.52 & 16.90 & 22.44 & 25.35 \\ \midrule
\multirow{16}{*}{\begin{tabular}[c]{@{}c@{}}Hanspell\\ (test)\end{tabular}} &
  \multirow{4}{*}{\klearner} &
  gleu &
  19.37 &
  15.23 &
  34.88 &
  0.00 &
  28.66 &
  22.64 &
  0.00 &
  23.84 &
  21.45 &
  18.91 &
  22.55 &
  20.55 &
  22.51 &
  21.78 &
  22.55 &
  30.36 \\
 &                           & prec      & 26.32 & 23.68 & 51.73 & 0.00   & 31.58 & 37.50 & 100.00 & 26.76 & 33.72 & 7.14  & 25.56 & 23.68 & 25.85 & 23.08 & 24.21 & 29.45 \\
 &                           & rec       & 2.60  & 3.09  & 18.00 & 0.00   & 4.26  & 4.62  & 100.00 & 4.58  & 5.06  & 1.19  & 4.56  & 3.14  & 3.32  & 2.78  & 4.48  & 5.33  \\
 &                           & $f_{0.5}$ & 9.32  & 10.16 & 37.63 & 0.00   & 13.82 & 15.46 & 100.00 & 13.59 & 15.80 & 3.57  & 13.30 & 10.26 & 10.97 & 9.39  & 12.87 & 15.46 \\
 & \multirow{4}{*}{\knative} & gleu      & 51.18 & 21.25 & 62.94 & 12.02  & 64.19 & 56.69 & 12.02  & 31.62 & 43.16 & 28.11 & 48.57 & 35.82 & 40.25 & 31.07 & 54.57 & 57.08 \\
 &                           & prec      & 72.97 & 69.23 & 75.38 & 88.89  & 92.31 & 85.28 & 50.00  & 57.06 & 63.86 & 81.82 & 78.01 & 66.06 & 77.50 & 68.42 & 79.82 & 81.93 \\
 &                           & rec       & 45.49 & 14.95 & 54.35 & 15.38  & 68.57 & 50.52 & 11.11  & 19.28 & 26.63 & 26.47 & 41.89 & 19.67 & 31.96 & 28.57 & 51.58 & 47.36 \\
 &                           & $f_{0.5}$ & 65.11 & 40.11 & 69.96 & 45.45  & 86.33 & 74.97 & 29.41  & 41.00 & 49.91 & 57.69 & 66.53 & 44.89 & 60.31 & 53.50 & 71.94 & 71.50 \\
 & \multirow{4}{*}{\klang}   & gleu      & 19.68 & 13.21 & 28.19 & 16.86  & 23.85 & 32.35 & 6.99   & 17.14 & 17.36 & 16.78 & 17.24 & 14.84 & 17.37 & 15.66 & 17.36 & 22.94 \\
 &                           & prec      & 32.13 & 33.07 & 45.79 & 29.54  & 29.09 & 59.42 & 24.00  & 28.75 & 26.57 & 28.17 & 28.20 & 23.83 & 26.11 & 29.26 & 27.49 & 29.18 \\
 &                           & rec       & 8.42  & 7.35  & 20.76 & 5.96   & 9.47  & 29.24 & 5.24   & 8.21  & 5.90  & 6.17  & 7.45  & 5.24  & 5.70  & 6.65  & 7.49  & 8.74  \\
 &                           & $f_{0.5}$ & 20.56 & 19.45 & 36.90 & 16.49  & 20.57 & 49.25 & 13.99  & 19.16 & 15.62 & 16.45 & 18.11 & 13.94 & 15.21 & 17.41 & 17.92 & 19.88 \\
 & \multirow{4}{*}{\kunion}  & gleu      & 21.89 & 13.57 & 31.73 & 16.81  & 31.06 & 46.91 & 7.51   & 19.67 & 19.94 & 18.31 & 19.79 & 17.14 & 19.17 & 17.41 & 21.15 & 28.82 \\
 &                           & prec      & 37.49 & 33.51 & 48.21 & 30.61  & 46.70 & 71.98 & 25.00  & 29.94 & 30.32 & 28.42 & 32.10 & 24.37 & 27.12 & 31.41 & 34.07 & 37.34 \\
 &                           & rec       & 10.10 & 7.40  & 21.25 & 6.14   & 15.48 & 39.30 & 5.26   & 8.14  & 6.62  & 6.12  & 8.60  & 5.01  & 5.64  & 7.08  & 9.69  & 11.58 \\
 &                           & $f_{0.5}$ & 24.31 & 19.64 & 38.46 & 17.03  & 33.28 & 61.72 & 14.29  & 19.50 & 17.67 & 16.43 & 20.76 & 13.75 & 15.40 & 18.61 & 22.66 & 25.85 \\ \midrule
\multirow{16}{*}{\begin{tabular}[c]{@{}c@{}}KoBART \\ + finetune \\ (valid)\end{tabular}} &
  \multirow{4}{*}{\klearner} &
  gleu &
  30.81 &
  29.15 &
  45.31 &
  0.00 &
  48.83 &
  35.07 &
  0.00 &
  38.95 &
  38.51 &
  37.23 &
  35.85 &
  42.99 &
  37.91 &
  32.06 &
  40.01 &
  46.94 \\
 &                           & prec      & 47.19 & 47.74 & 46.12 & 100.00 & 48.60 & 21.57 & 100.00 & 42.50 & 44.37 & 46.92 & 41.95 & 49.38 & 46.45 & 43.99 & 45.07 & 43.95 \\
 &                           & rec       & 19.11 & 24.09 & 28.46 & 100.00 & 31.16 & 8.53  & 100.00 & 21.98 & 21.58 & 26.79 & 21.95 & 28.12 & 23.92 & 21.07 & 25.19 & 26.35 \\
 &                           & $f_{0.5}$ & 36.47 & 39.89 & 41.02 & 100.00 & 43.68 & 16.45 & 100.00 & 35.81 & 36.63 & 40.76 & 35.48 & 42.89 & 39.08 & 36.13 & 38.92 & 38.76 \\
 & \multirow{4}{*}{\knative} & gleu      & 61.12 & 43.40 & 51.25 & 57.13  & 66.49 & 71.33 & 20.20  & 46.43 & 57.91 & 43.30 & 58.06 & 57.61 & 49.85 & 57.19 & 65.11 & 69.37 \\
 &                           & prec      & 74.03 & 69.51 & 65.32 & 65.99  & 85.13 & 79.25 & 61.11  & 69.95 & 73.05 & 70.21 & 66.51 & 73.68 & 68.61 & 77.14 & 76.50 & 75.07 \\
 &                           & rec       & 51.09 & 37.37 & 38.90 & 47.55  & 59.70 & 62.68 & 33.33  & 34.64 & 40.87 & 37.04 & 46.89 & 39.51 & 38.40 & 53.87 & 57.99 & 56.81 \\
 &                           & $f_{0.5}$ & 67.92 & 59.31 & 57.50 & 61.24  & 78.44 & 75.27 & 52.19  & 58.08 & 63.09 & 59.52 & 61.36 & 62.78 & 59.28 & 70.99 & 71.90 & 70.53 \\
 & \multirow{4}{*}{\klang}   & gleu      & 21.36 & 18.22 & 31.65 & 18.81  & 25.35 & 23.19 & 16.61  & 21.89 & 25.20 & 24.62 & 23.19 & 22.90 & 23.82 & 20.10 & 22.50 & 28.57 \\
 &                           & prec      & 35.88 & 45.60 & 48.41 & 37.63  & 45.81 & 41.98 & 38.18  & 39.35 & 41.99 & 45.16 & 38.59 & 44.46 & 42.35 & 39.19 & 35.20 & 37.69 \\
 &                           & rec       & 8.70  & 12.97 & 21.48 & 9.36   & 15.25 & 13.14 & 10.17  & 11.91 & 13.04 & 14.44 & 11.93 & 13.20 & 12.53 & 10.36 & 10.43 & 12.64 \\
 &                           & $f_{0.5}$ & 22.05 & 30.30 & 38.69 & 23.46  & 32.63 & 29.15 & 24.61  & 26.91 & 29.04 & 31.65 & 26.64 & 30.12 & 28.67 & 25.15 & 23.84 & 26.96 \\
 & \multirow{4}{*}{\kunion}  & gleu      & 23.59 & 18.64 & 36.52 & 21.15  & 32.08 & 36.08 & 15.50  & 26.60 & 29.25 & 26.64 & 25.48 & 27.57 & 26.69 & 22.59 & 26.62 & 34.07 \\
 &                           & prec      & 40.52 & 46.03 & 52.46 & 44.78  & 51.94 & 60.81 & 33.58  & 41.89 & 45.21 & 49.39 & 41.50 & 47.83 & 46.44 & 45.55 & 41.35 & 44.33 \\
 &                           & rec       & 10.10 & 12.50 & 24.34 & 11.85  & 15.97 & 21.91 & 8.27   & 12.74 & 13.67 & 15.83 & 13.18 & 14.48 & 13.55 & 12.97 & 13.21 & 15.33 \\
 &                           & $f_{0.5}$ & 25.27 & 29.92 & 42.60 & 28.77  & 35.74 & 44.79 & 20.83  & 28.72 & 30.90 & 34.67 & 29.00 & 32.71 & 31.24 & 30.27 & 28.97 & 32.13 \\ \midrule
\multirow{16}{*}{\begin{tabular}[c]{@{}c@{}}KoBART \\ + finetune \\ (test)\end{tabular}} &
  \multirow{4}{*}{\klearner} &
  gleu &
  31.05 &
  25.37 &
  42.99 &
  0.00 &
  41.12 &
  25.49 &
  0.00 &
  35.83 &
  35.38 &
  35.25 &
  35.78 &
  40.53 &
  38.22 &
  35.44 &
  38.96 &
  45.06 \\
 &                           & prec      & 41.60 & 43.17 & 45.89 & 0.00   & 51.96 & 39.43 & 100.00 & 42.56 & 44.24 & 37.54 & 40.11 & 48.88 & 48.18 & 48.20 & 42.49 & 43.35 \\
 &                           & rec       & 15.23 & 20.10 & 27.00 & 0.00   & 23.64 & 14.36 & 100.00 & 19.89 & 21.36 & 19.45 & 19.89 & 26.32 & 23.56 & 25.29 & 23.44 & 24.54 \\
 &                           & $f_{0.5}$ & 30.89 & 35.09 & 40.25 & 0.00   & 41.86 & 29.22 & 100.00 & 34.65 & 36.43 & 31.64 & 33.33 & 41.71 & 39.84 & 40.80 & 36.54 & 37.58 \\
 & \multirow{4}{*}{\knative} & gleu      & 57.71 & 46.07 & 56.02 & 39.23  & 69.69 & 67.94 & 42.97  & 45.61 & 56.40 & 59.54 & 59.45 & 59.61 & 47.03 & 43.38 & 64.57 & 67.24 \\
 &                           & prec      & 74.75 & 73.48 & 61.14 & 71.10  & 85.38 & 79.45 & 65.97  & 69.15 & 73.66 & 76.32 & 68.68 & 77.53 & 72.41 & 67.50 & 77.70 & 75.34 \\
 &                           & rec       & 48.73 & 40.42 & 41.54 & 41.02  & 66.67 & 60.17 & 42.59  & 34.06 & 40.70 & 47.06 & 48.77 & 44.63 & 36.25 & 41.03 & 58.03 & 55.95 \\
 &                           & $f_{0.5}$ & 67.53 & 63.13 & 55.84 & 61.97  & 80.81 & 74.65 & 59.37  & 57.32 & 63.39 & 67.81 & 63.49 & 67.54 & 60.33 & 59.78 & 72.77 & 70.45 \\
 & \multirow{4}{*}{\klang}   & gleu      & 22.29 & 17.64 & 31.13 & 20.61  & 31.14 & 21.49 & 12.42  & 21.67 & 24.87 & 26.36 & 23.31 & 22.39 & 24.62 & 20.95 & 22.55 & 28.48 \\
 &                           & prec      & 36.23 & 45.31 & 45.98 & 39.63  & 49.45 & 44.27 & 29.64  & 39.95 & 41.03 & 47.55 & 38.21 & 44.18 & 42.71 & 40.07 & 36.67 & 37.56 \\
 &                           & rec       & 8.12  & 12.14 & 18.77 & 9.13   & 17.26 & 12.73 & 6.99   & 10.85 & 11.50 & 13.01 & 10.92 & 11.95 & 11.56 & 10.20 & 10.07 & 11.62 \\
 &                           & $f_{0.5}$ & 21.36 & 29.25 & 35.61 & 23.72  & 35.96 & 29.58 & 17.96  & 25.97 & 27.05 & 31.01 & 25.43 & 28.64 & 27.71 & 25.23 & 23.95 & 25.93 \\
 & \multirow{4}{*}{\kunion}  & gleu      & 23.67 & 18.22 & 36.10 & 20.29  & 33.42 & 34.83 & 13.76  & 25.09 & 27.96 & 29.27 & 25.95 & 27.15 & 27.72 & 23.68 & 26.44 & 33.70 \\
 &                           & prec      & 40.58 & 45.64 & 52.16 & 42.35  & 58.52 & 63.26 & 37.89  & 43.76 & 46.45 & 49.16 & 41.30 & 49.35 & 47.51 & 46.67 & 42.69 & 44.75 \\
 &                           & rec       & 9.47  & 12.05 & 22.54 & 10.06  & 19.61 & 21.48 & 9.04   & 12.46 & 13.29 & 13.92 & 12.34 & 14.30 & 13.23 & 12.70 & 12.68 & 14.64 \\
 &                           & $f_{0.5}$ & 24.47 & 29.28 & 41.30 & 25.78  & 41.84 & 45.48 & 23.11  & 29.11 & 30.98 & 32.61 & 28.09 & 33.10 & 31.28 & 30.38 & 28.97 & 31.70 \\ \midrule
\multirow{12}{*}{\begin{tabular}[c]{@{}c@{}}KoBART \\ + union \\ +finetune\\ (valid)\end{tabular}} &
  \multirow{4}{*}{\klearner} &
  gleu &
  27.37 &
  28.19 &
  45.76 &
  0.00 &
  41.25 &
  35.35 &
  0.00 &
  38.06 &
  35.58 &
  36.79 &
  33.46 &
  38.87 &
  35.67 &
  28.91 &
  37.38 &
  44.66 \\
 &                           & prec      & 50.01 & 55.82 & 57.04 & 100.00 & 55.76 & 41.75 & 100.00 & 52.56 & 49.49 & 58.63 & 49.25 & 55.87 & 55.75 & 54.11 & 50.97 & 51.94 \\
 &                           & rec       & 14.93 & 20.69 & 30.22 & 100.00 & 22.22 & 13.18 & 100.00 & 21.13 & 18.92 & 26.03 & 20.07 & 24.04 & 21.42 & 20.27 & 22.40 & 23.55 \\
 &                           & $f_{0.5}$ & 33.98 & 41.64 & 48.42 & 100.00 & 42.79 & 29.05 & 100.00 & 40.49 & 37.40 & 46.86 & 38.14 & 44.16 & 42.19 & 40.53 & 40.59 & 41.83 \\
 & \multirow{4}{*}{\knative} & gleu      & 52.63 & 34.43 & 50.01 & 44.80  & 50.88 & 62.02 & 19.93  & 39.88 & 51.19 & 41.17 & 49.43 & 51.92 & 44.72 & 53.01 & 58.43 & 61.64 \\
 &                           & prec      & 76.54 & 74.25 & 75.03 & 69.65  & 90.85 & 88.28 & 100.00 & 73.44 & 83.64 & 74.32 & 75.39 & 77.49 & 83.49 & 92.65 & 80.50 & 83.95 \\
 &                           & rec       & 42.21 & 25.88 & 39.75 & 41.67  & 49.26 & 52.94 & 33.33  & 25.29 & 35.32 & 33.33 & 39.61 & 32.57 & 34.78 & 48.21 & 51.52 & 48.55 \\
 &                           & $f_{0.5}$ & 65.79 & 54.05 & 63.71 & 61.39  & 77.68 & 77.88 & 71.43  & 53.19 & 65.61 & 59.49 & 63.84 & 60.73 & 65.22 & 78.16 & 72.35 & 73.25 \\
 & \multirow{4}{*}{\klang}   & gleu      & 21.40 & 17.87 & 32.12 & 19.70  & 23.76 & 24.02 & 15.11  & 22.28 & 25.37 & 25.51 & 22.91 & 22.76 & 22.95 & 20.73 & 22.31 & 28.51 \\
 &                           & prec      & 36.78 & 45.27 & 50.17 & 39.98  & 44.92 & 44.67 & 37.59  & 40.16 & 42.00 & 46.86 & 38.37 & 44.85 & 42.43 & 42.98 & 35.47 & 38.53 \\
 &                           & rec       & 8.79  & 12.58 & 22.38 & 9.97   & 13.71 & 14.50 & 8.67   & 12.08 & 12.49 & 14.83 & 11.55 & 12.81 & 11.86 & 12.05 & 10.45 & 12.72 \\
 &                           & $f_{0.5}$ & 22.46 & 29.79 & 40.19 & 24.94  & 30.86 & 31.53 & 22.54  & 27.41 & 28.52 & 32.72 & 26.20 & 29.89 & 28.00 & 28.39 & 23.98 & 27.40 \\ \midrule
\multirow{12}{*}{\begin{tabular}[c]{@{}c@{}}KoBART \\ + union \\ +finetune\\ (test)\end{tabular}} &
  \multirow{4}{*}{\klearner} &
  gleu &
  28.45 &
  22.88 &
  43.09 &
  0.00 &
  35.61 &
  25.51 &
  0.00 &
  32.85 &
  33.28 &
  29.59 &
  32.73 &
  37.01 &
  34.93 &
  33.36 &
  35.74 &
  42.66 \\
 &                           & prec      & 47.90 & 47.19 & 59.30 & 0.00   & 64.04 & 53.56 & 100.00 & 50.13 & 53.62 & 49.76 & 47.99 & 58.27 & 57.13 & 56.74 & 51.51 & 53.51 \\
 &                           & rec       & 12.18 & 14.61 & 29.07 & 0.00   & 16.31 & 13.33 & 100.00 & 16.98 & 18.80 & 15.48 & 17.25 & 21.82 & 19.64 & 20.49 & 20.83 & 21.18 \\
 &                           & $f_{0.5}$ & 30.19 & 32.62 & 49.08 & 0.00   & 40.36 & 33.21 & 100.00 & 36.05 & 39.12 & 34.44 & 35.37 & 43.67 & 41.34 & 41.91 & 39.80 & 41.00 \\
 & \multirow{4}{*}{\knative} & gleu      & 47.54 & 37.94 & 56.58 & 30.31  & 62.21 & 58.44 & 32.32  & 38.89 & 49.33 & 58.46 & 50.69 & 54.55 & 44.65 & 41.05 & 56.89 & 59.71 \\
 &                           & prec      & 78.05 & 81.52 & 81.58 & 77.16  & 93.37 & 88.67 & 69.84  & 73.00 & 83.80 & 89.15 & 76.02 & 85.26 & 85.53 & 78.91 & 82.82 & 85.47 \\
 &                           & rec       & 38.47 & 29.79 & 47.22 & 39.10  & 60.48 & 49.78 & 29.63  & 25.45 & 34.67 & 48.04 & 39.16 & 37.43 & 32.64 & 41.39 & 50.80 & 47.38 \\
 &                           & $f_{0.5}$ & 64.68 & 60.50 & 71.15 & 64.56  & 84.17 & 76.68 & 54.75  & 53.14 & 65.28 & 76.04 & 63.95 & 67.89 & 64.57 & 66.77 & 73.54 & 73.63 \\
 & \multirow{4}{*}{\klang}   & gleu      & 21.94 & 17.43 & 32.56 & 20.44  & 31.91 & 22.52 & 12.10  & 21.92 & 24.55 & 26.88 & 23.48 & 22.35 & 24.73 & 21.33 & 22.43 & 28.65 \\
 &                           & prec      & 36.55 & 44.07 & 47.59 & 40.51  & 49.46 & 45.42 & 28.62  & 40.61 & 40.65 & 47.45 & 38.01 & 43.70 & 42.04 & 42.22 & 36.49 & 37.46 \\
 &                           & rec       & 8.35  & 11.89 & 20.52 & 10.10  & 18.44 & 14.24 & 7.28   & 11.47 & 11.25 & 13.27 & 11.20 & 12.10 & 11.49 & 11.31 & 10.22 & 12.00 \\
 &                           & $f_{0.5}$ & 21.81 & 28.56 & 37.64 & 25.26  & 36.93 & 31.55 & 18.00  & 26.90 & 26.68 & 31.28 & 25.68 & 28.68 & 27.42 & 27.26 & 24.08 & 26.28 \\
 \bottomrule

\caption{Full scores on all error types, all datasets, on all methods, including valid dataset and test dataset. We also provide individual dataset count by error types on the top.}\\
\label{tabs:errortype_scores_on_everything}
\end{longtable}
\twocolumn
\endgroup

\end{appendix}

\end{document}